# Advancements in Crop Analysis through Deep Learning and Explainable AI

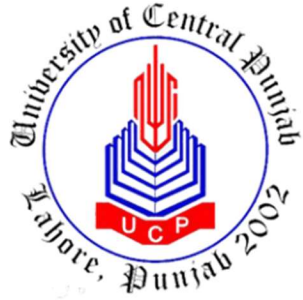

MASTER OF SCIENCE
IN
DATA SCIENCE

Submitted By
Hamza Khan
L1S22MSDS0011

DEPARTMENT OF COMPUTER SCIENCES
FACULTY OF INFORMATION TECHNOLOGY & COMPUTER SCIENCES
UNIVERSITY OF CENTRAL PUNJAB

July 2025

# Advancements in Crop Analysis through Deep Learning and Explainable AI

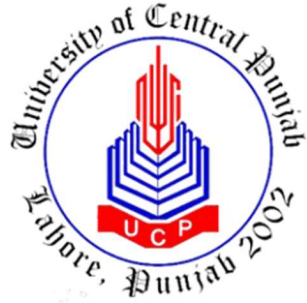

A Thesis submitted in partial fulfillment
of the requirements for the degree of

MASTER OF SCIENCE
IN
DATA SCIENCE

Submitted By
Hamza Khan
L1S22MSDS0011

Supervised By
Dr. Rabia Tehseen

DEPARTMENT OF COMPUTER SCIENCES
FACULTY OF INFORMATION TECHNOLOGY & COMPUTER SCIENCES
UNIVERSITY OF CENTRAL PUNJAB

July 2025

# ABSTRACT

Rice is a staple food of world significance in terms of being a major international trade, economic growth, and nutrition. Among the Asian countries that made significant contributions into the sphere of rice growing and rice consumption, one may distinguish China, India, Pakistan, Thailand, Vietnam, and Indonesia. These are the countries that are famous in the production of the long as well as the short rice grains. To fit various gastronomic palates and some cultural practices, these sizes are further classified into arborio, ipsala, kainat saila, jasmine and basmati among others. In order to meet consumer needs and alter a reputation of a country, it is important to monitor the crops of rice, as well as estimate the quality of rice grain. Since the manual inspection is time consuming, prone to mistakes and labour-intensive, an automatized solution should be adopted to assure quality control in the manufacturing phase and enhance the yield of farmers.

To start with, we have made a proposal of automatic processing of different types of rice grains (five of them), and it is based on the convolutional neural network (CNN) involving the explained AI methods. The testing and training was performed on a publically available dataset of 75,000 images across 15K images of each type of rice grain. To evaluate the proposed model, we used the performance criteria measured with the help of special values such as accuracy, recall, precision, and F1-Score. Following the extensive training and validation, the CNN model presented high accuracy rate and outstanding area under the Receiver Operating Characteristic (ROC) curve in every class. The confusion matrix supported this claim as there were small misclassifications in the various rice varieties which indicated that the different and distinct varieties were identified using the model.

Secondly, an accurate and effective method of diagnosing rice leaf disease namely Brown Spot, Blast, Bacterial Blight and Tungro was presented. The whole apparatus was divided into two parts: explainable AI in the first part and CNN, VGG16, RESNET-50, and MobileNETV2 in the second.

Moreover, introduction of the explainability methods such as SHAP (SHapley Additive exPlanations) and LIME (Local Interpretable Model-agnostic Explanations) provided valuable data regarding the mechanism of decision-making within the model, and was able to show how the specific features of the rice grains influenced the outcome of the classification. This interpretability makes the model more applicable to real life scenario and confidence on model predictions is raised. The findings create a path towards better automated classification systems because it proved the great potential of deep learning methods in agriculture.

# DEDICATION

"To my father, ***Abdul Jabbar Khan (Late)***,

My Mother,

and

My Family

# ACKNOWLEDGEMENTS

- All Praises to **GOD Almighty** by whose Grace I was able to complete this research project. Thanks to HIM, who blessed me with the best at every step of my life.

- I would like to express my deepest gratitude and appreciation to **Prophet Muhammad (S.A.W)** for his profound teachings and exemplary life, which have served as a guiding light and source of inspiration throughout my thesis journey.

- We wish to offer our thanks to **Dr. Amjad Iqbal, Dean, FoIT&CS, UCP** who has been the main icon in our gearing up towards accomplishing something fruitful and helpful for the amelioration of society around us.

- I am deeply grateful to my research supervisor **Dr. Rabia Tehseen, Assistant Professor, FoIT&CS, UCP** for her constant support, and knowledge throughout the thesis process. Her leadership and effort have helped shape the direction and success of this research project. Her valuable comments, suggestions, and generous criticism helped me in writing this Thesis. I am very grateful to her from the core of my heart for her expert guidance and sympathetic attitude. Their mentorship and guidance have been pivotal in my academic and research journey, and I am truly fortunate to have had the opportunity to work under their supervision.

- I would also like to thank **Mr. Saad** for teaching us Tools and Techniques in Data Science in such a way that he provided us a solid foundation of different techniques of data science. I would also like to Thank **Dr. Shazia Saqib** for teaching us the course of **Deep Learning** by adopting the practical approach.

- Furthermore, I would like to extend my gratitude to my classmate, **Engr. Muhammad Junaid, Assistant Manager (Tech) at Artificial Intelligence Technology Centre**

**(AITeC), National Centre for Physics (NCP),** for his invaluable guidance and unwavering support throughout this process.

# DECLARATION

I, *Hamza Khan* S/O *Abdul Jabbar Khan*, a student of *"Master of Science in Data Science"*, at **"Faculty of Information Technology & Computer Sciences"**, **University of Central** Punjab (UCP), hereby declare that this thesis titled, *"Advancements in Crop Analysis through Deep Learning and Explainable AI"* is my own research work and has not been submitted, published, or printed elsewhere in Pakistan or abroad. Additionally, I will not use this thesis to obtain any degree other than the one stated above. I fully understand that if my statement is found to be incorrect at any stage, including after the award of the degree, the University has the right to revoke my MS/M.Phil. degree.

**Signature of Student:**

**Name of Student:** Hamza Khan

**Registration Number:** L1S22MSDS0011

**Date:**

# PLAGIARISM UNDERTAKING

I solemnly declare that the research work presented in this thesis titled, ***"Advancements in Crop Analysis through Deep Learning and Explainable AI"*** is solely my research work, and that the entire thesis has been completed by me, with no significant contribution from any other person or institution. Any small contribution, wherever taken, has been duly acknowledged.

I understand the zero-tolerance policy of the HEC and University of Central Punjab towards plagiarism. Therefore, I as an author of the above titled thesis declare that no portion of my thesis has been plagiarized and that every material used by other sources has been properly acknowledged, cited, and referenced.

I undertake that if I am found guilty of any formal plagiarism in the above titled thesis, even after the award of MS/MPhil. degree, the University reserves the right to revoke my degree, and that HEC and the University have the right to publish my name on the HEC/University website for submitting a plagiarized thesis.

**Signature of Student:**

**Name of Student:** Hamza Khan

**Registration Number:** L1S22MSDS0011

**Date:**

# CERTIFICATE OF RESEARCH COMPLETION

It is to certify that thesis titled, **"Advancements in Crop Analysis through Deep Learning and Explainable AI"**, submitted by **Hamza Khan,** Registration No. **L1S22MSDS0011**, for MS degree at **"Faculty of Information Technology and Computer Sciences", University of Central Punjab (UCP)**, is an original research work and contains satisfactory material to be eligible for evaluation by the Examiner(s) for the award of the above stated degree.

**Dr. Rabia Tehseen**
Assistant Professor
Faculty of IT and CS
University of Central Punjab

                                                    Signature

Date: \_\_\_\_\_\_\_\_\_\_\_\_\_\_\_\_\_\_

# CERTIFICATE OF EXAMINERS

It is certified that the research work contained in this thesis titled **"Advancements in Crop Analysis through Deep Learning and Explainable AI"** is up to the mark for the award of **"Master of Science in Data Science".**

**Internal Examiner**

**Signature:**

**Name:** ______________________

**Date:** ______________________

**External Examiner:**

**Signature:**

**Name:** ______________________

**Date:** ______________________

**Dean**
Faculty of IT & CS
University of Central Punjab (UCP)

**Signature:**

**Name:**   Dr. Muhammad Amjad Iqbal

**Date:** ______________________

# TABLE OF CONTENTS

# LIST OF FIGURES

# LIST OF TABLES

# LIST OF ABBREVIATIONS AND ACRONYM

| | |
|---|---|
| **UAV** | Unmanned Aerial Vehicles |
| **VGG** | Virtual Geometry Group |
| **CNN** | Convolutional Neural Network |
| **ANN** | Artificial Neural Network |
| **KNN** | k-Nearest Neighbor |
| **DNN** | Deep Neural Network |
| **LR** | Logistic Regression |
| **NB** | Naïve Bayes |
| **SVM** | Support Vector Machines |
| **GA** | Genetic Algorithm |
| **UN** | United Nations |
| **RESNET** | Residual Network |
| **xAI** | Explainable AI |
| **LIME** | Local Interpretable Model-agnostic Explanations |
| **SHAP** | SHapley Additive exPlanations |
| **ROC** | Receiver Operating Characteristic |
| **AUC** | Area under the curve |
| **GLCM** | Gray-Level Co-Occurrence Matrix |
| **FPR** | False to Positive Ratio |
| **FNR** | False to Negative Ratio |

# CHAPTER ONE: INTRODUCTION

## 1.1 Crop Analysis

Crop analysis, which is the most important aspect employed in the contemporary precision agriculture, is a scientific study of different aspects of crop production and management with an objective of maximizing yield, quality, supply chain, and sustainability by employing the latest advancements in deep learning and computer vision approaches. It involves collection and assessment of information on various issues such as the condition of soil, environmental detection, condition of plant, pest and disease condition, method of irrigation as well as nutrient intensity [1]. The main purpose of crop analysis is to obtain significant information on crop health and crop growth performance in order to make decisions on agricultural practice with knowledge [2]. It obtains numerous kinds of valuable information, including the estimation of the yield, crop identification, growth stage, and health condition, using the wide range of data sources, including the use of satellites, unmanned aerial vehicles or ground-mounted cameras to collect the needed data [3], [4].

The upgrade of crop quality and productivity is among the primary benefits of crop analysis. By keeping a close track of several parameters, including plant growth, nutrient levels, and soil health among others, farmers can make informed choices that lead directly to improved yield and an overall better quality of their produce [5]. Also, the crop analysis reduces expenditures by making the most efficient member of resources required, such as pesticides, fertilizers, and water. Targeted resource application according to analytical data will guarantee effective placement of local resources and reduce resources as well as costs [6].

Besides, as a way of encouraging a sustainable form of travelling, appropriate management of the resources and prevention of overuse of the chemicals, part of crop analysis contributes to ecologically friendly and sustainable approaches to agriculture, thereby keeping the ecosystem healthy and keeping the soil healthy to reap the crop again [7], [8]. Moreover, it also enables the detection of such issues as an outbreak of disease, lack of nutrients, or pest invasion at the early stages and prevents the further spreading of the problem to reduce losses and the destruction of the crops on a large scale [5], [9].

Last but not least, crop analysis also enables precision agriculture, a procedure of information-driven farming that uses analytics and contemporary technology to manage each portion of crop creation. Proper and timely information helps farmers to make correct and informed decisions on planting, irrigation, fertilization, and harvesting activities that can make agricultural practices effective, profitable and sustainable [2], [5].

Based on applications, our research thesis will base its focus on two major applications of crop analysis, namely crop disease detection and crop classification. ***The method or steps to define the various kinds of crops properly can be termed as crop classification and the way of defining the products or commodities that are grown using these crops can be referred as product classification*** [10].

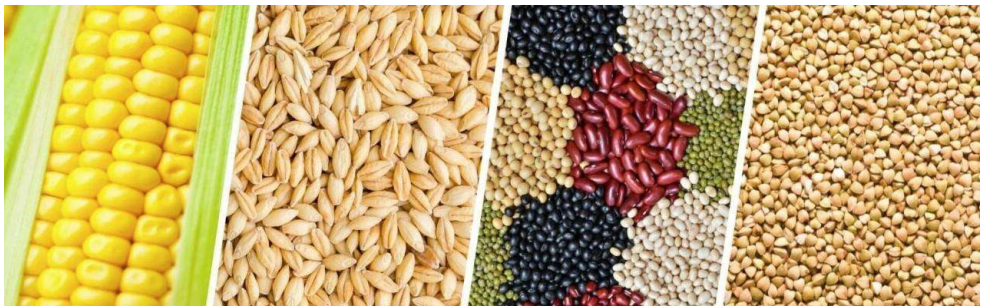

**Figure 1.1: A general picture depicting different types of products generated from crops**

*Crop disease detection, on the other hand, is a meticulous procedure that involves monitoring plant health, identifying symptoms, identifying pathogens or stress factors, and diagnosing diseases affecting crops using a variety of techniques and technologies* [11].

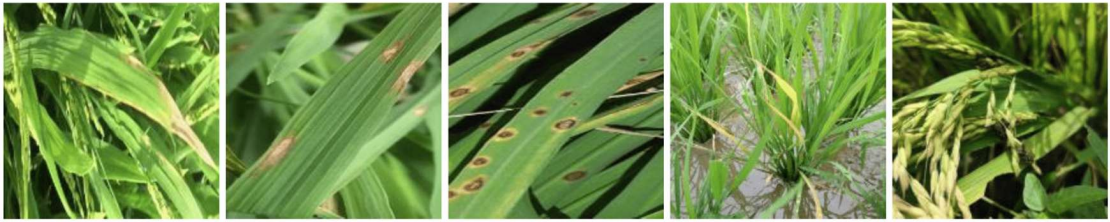

**Figure 1.2: A general picture depicting crops with different diseases**

Crop disease detection involves different kinds of tasks such as infections caused by various microorganisms including fungi, bacteria, viruses, or pests. These can be analyzed and detected by leveraging the uses of advanced deep learning techniques with combination of image processing and computer vision [11].

## 1.2 Motivation:

According to the United Nations (UN), the global population has been progressively rising. In 2022, the world population reached 8 billion. Projections suggest that it will reach 9 billion by 2037 and 10 billion by the conclusion of 2060. Remarkably, the population doubled from 3 billion in 1959 to 6 billion in 1999 within a span of just 40 years [12]. [13], [14], [15].

As of 2025, the global population is expanding at a rate of approximately 0.85% per year, resulting in an addition of around 70 million individuals annually. The growth rate reached its peak in the late 1960s at 2.09% but has been on a downward trend since then. It is estimated that at 2047 this rate is expected to further drop to a value less than 0.50% and will attain a value of zero in the year 2084 and will still fall to a negative value of -0.12 percent by the year 2100 [16], [17].

The population is on the rise which increases the demand for food. New ideas have to be introduced to ensure that the prevention of hunger will be through sustainable methods. Through this ever-increasing population, agriculture faces major challenges, especially in avenues like the accurate classification of crops and timely identification of crop diseases. Crop variety recognition is very crucial, and this is the only way towards getting the best results in terms of harvest and securing sources of food against hunger. It helps the farmers to maximize production and reduce losses by adopting some specific techniques of management [18], [19], [20].

On the other hand, detecting diseases at early parts and when its necessary is also beneficial because it can help minimize the effects of disease on the crops. The timely diagnosis of disease would also be useful in preventing infection development by using rapid treatment and minimizing the use of chemical drugs. The farmers can reduce the impact of the disease by emphasizing the infected regions, and by the application of the customized remedies, such as the application of an appropriate fungicide and organizing the irrigation regime. Besides helping to preserve crops, this practice will also encourage sustainable forms of agriculture as less chemicals will be used in the cultivation process and the species that are beneficial to the ecosystem will be maintained. A stable and well organized disease management system in the long run is quite vital on achieving the food security and restoring the agricultural systems [21], [22].

We have been motivated by this and have built upon this idea by having a combined framework that has been using state of the art technologies such as computer vision and deep learning in order to correctly identify different types of rice grain and diagnosing the disease of rice crops. Rice grain differentiation is proposed to use an effective and precise classification method in order to implement maximum best management practices and improve total yield. The suggested framework also covers a disease detection system based on the visual data analysis that allows detecting the rice diseases in an early stage and allows treating them on time. Such a twofold approach encourages sustainable farming practices by ensuring conservation of helpful species and assisting farmers to employ less chemicals. This would enhance sustainability in the agricultural sector, a healthy environment, and food security in the long-term perspective as this will reduce both the classification and disease problems as they arise. These systems are being integrated to encourage a comprehensive way of rice production, which ensures that farmers are also able to manage their rice sufficiently and to minimise their impact on the ecency.

## 1.1 Rice Grain Classification:

With the recent advances in the domain of AI, researchers and students from the field of agriculture focuses to integrate the AI in to crop analysis by using the combined applications of deep learning, machine learning and computer vision. This integration can be useful to execute various tasks in the field of agriculture such as monitoring of crop health [23], [24], [25] yield estimation [26], [27], [28], [29], price prediction [30], [31], [32] optimized usage of pesticides [33] and fertilizer application [34]. Crop classification is a fundamental component of crop analysis, involves categorizing the different varieties of crops. Efficiency of different agricultural tasks such as predictive modeling, well-informed decision making, and advancements of sustainable farming practices can be improved using precise categorization of crop varieties [35].

As defined earlier, crop classification refers to classification of different crops in field, while the product classification can be referred as the categorization of different products generated from those crops. In the first part of our research, we proposed an efficient and automatic framework for categorization of different varieties of rice grains. This categorization can be useful for farmers as well as industrialists to ensure the accuracy of quality and supply chain processes.

It is a time-consuming and laborious process to classify the different varieties of rice grain manually by conducting detailed assessments of different morphological features such as size, shape and color etc. For large-scale agricultural setups, this labor-intensive procedure can be expensive and ineffective. By quickly analyzing large datasets according to preset criteria, automated classification systems that combine deep learning and computer vision can offer a workable solution. By using this method, we could create a classification system that is more accurate, saving money, time, and reducing human error [36]. The efficacy of the study can be evaluated utilising an array of evaluation metrics, including F1-Score, accuracy, precision, and recall [37].

One of the most important of the various staple crops is rice, which is grown in more than 100 countries and is produced in large quantities in China and Southeast Asia. It supplies food resources to over 3.5 billion people worldwide and ranks third in cultivation, behind maize and sugarcane. It is also a major source of income for 200 million households in developing nations because of its low cost, ease of preparation, and extended shelf life [8], [38], [39].

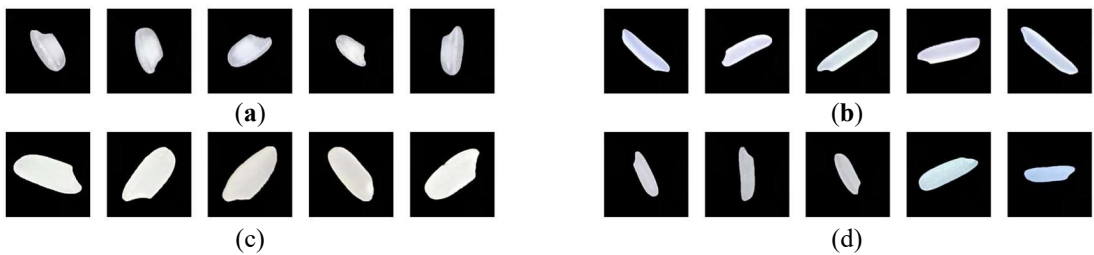

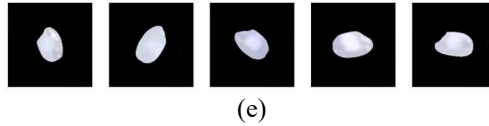
(e)

**Figure 1.3: Different Varieties of Rice grain (a) Arborio Rice Grains (b) Basmati Rice Grains (c) Ipsala Rice Grains (d) Jasmine Rice Grains (e) Karacadag Rice Grains**

In this study, five different rice grain varieties are categorized using an automatic and effective classification system that takes into account their morphological characteristics, including size, shape, color, eccentricity, perimeter, major axis length, and minor axis length. Out of the five types of grains, we have selected Arborio, Basmati, Ipsala, Jasmine, and Karacadag grains for classification (as shown in *Figure 1.3*).

For image processing based deep learning tasks, Convolutional Neural Networks (CNN) are the most efficient and useful approach to handle large volume of data. CNNs works in raw input images by learning the hierarchal features, resulting in high accuracy. It helps researchers and scientists to analyze and classify different varieties of rice grains, results in well-informed decision making,

Convolutional Neural Networks (CNN) are robust deep learning models renowned for their proficiency in processing and analysis of image data. Raw image data is used for learning of hierarchal features by leveraging the use of CNN, which results in accurate classification for different varieties of rice grains, to help researchers and farmers for enhanced and well-informed decision making in harvesting, supply chain and other purposes [40], [41]. CNN can identify numerous patterns, and properties in images of rice grains, including grain size, shape, color, and other morphological features. This capability is particularly helpful in the agricultural environment.

A lot of research has been carried out by leveraging the use of different CNN architectures such as VGG16, VGG19, and MobileNetV3-Small have been proposed for different tasks related to crop analysis. Regardless of this matter, there is a limitation of these models to understand the explain-ability of model decisions and their applicability in real world scenarios. There is a need to integrate deep learning models with xAI approaches to ensure transparency, resulting in improved accuracy and efficiency in classification processes. This will ultimately benefit quality analysis for farmers and the agricultural industry.

A framework based on a combination of different architectures of CNN and XAI approaches is presented in our research work. Various CNN architectures, including RESNET-50, MobileNet-V2, InceptionNet, and VGG16, are employed to extract different morphological characteristics of rice grains. For optimal decision transparency, these suggested deep learning models are then combined with two distinct XAI techniques: Shapley Additive explanations and Local Interpretable Model-agnostic Explanations (LIME). To increase the interpretability of CNN models in agricultural practices, it is imperative to incorporate these XAI techniques. While SHAP offers a unified measure of feature importance based on game theory for consistent explanations, LIME helps identify important features in predictions. This increases the model's agricultural efficacy.

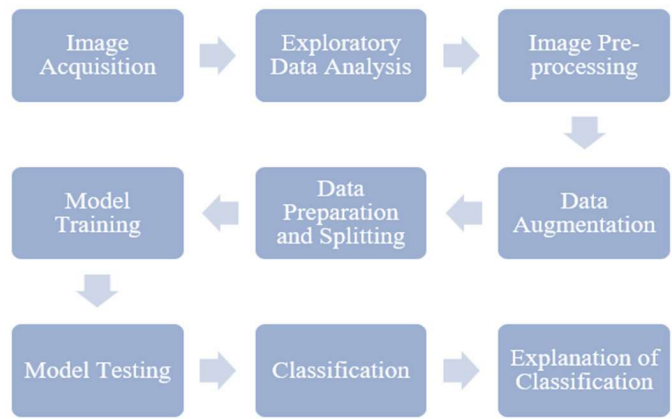

**Figure 1.4: Block Diagram of Proposed Rice Grain Classification System**

Convolutional neural networks (CNN) are used to classify images into the appropriate categories. Shapley Additive Explanations (SHAP) and Local Interpretable Model-agnostic Explanations (LIME) are then used to create interpretable models. By improving classification accuracy and providing insights into the model's decision-making process, this two-step process enhances the model's overall performance. The block diagram of our proposed system is *(as shown in Figure 1.4).*

## 1.2 Rice Crop Disease Detection:

Any mutation that changes the normal plant behaviors is termed as a plant disease, which also affects the amount of crop productions negatively, since their quality and quantity reduce. Effective detection of crop disease is a step by step procedure of determining and diagnosing the different crop diseases that may attack crops with different methods and equipment. It is a very scrupulous procedure, which entails monitoring of the plant's health, observation of any visible symptoms, and determining causative pathogens or stressors of the diseases. Due to the increase in the population of the world, crop disease detection is growing in importance to guarantee food security, the practice of sustainable farming, and the amplification of agricultural production [40].

Management of large-scale crop yields involves rapid responses, which include close observation of diseases and incorporation of short-term solutions. The diseases of crops have the capacity to greatly reduce the capacity of a plant leaving it with weaker rates of growth, low fruit yield, premature fall of its leaves, and other types of sicknesses. These diseases are most often caused by pathogens such as bacteria, viruses, or fungi and may spread with a high rate of speed between crops. By means of the seeds too they can disseminate locations. Hence, they endanger the health and productivity of crops and are particularly dangerous, as early diagnosis and treatment are needed to ensure their health [42], [43], [44].

Crop diseases may trouble their plant stems, leaves, vegetables and fruits and they are caused by bacterial, fungal, viral and even insect pests infection. The usual stages of the detection procedure are locating the places where there are afflicted regions, recording the distinctive personalities of the diseases, and categorizing them accordingly. The process of crop disease identification has traditionally been a cumbersome experience that is required to involve the expertise of plant pathologists and agronomists. Nevertheless, finding and talking to such experts can often be expensive, time-consuming, cumbersome, and particularly so on the case of the large agricultural areas. The problem is even more challenging because there is a necessity to satisfy the need to respond quickly to outbreaks because they may cause big losses.

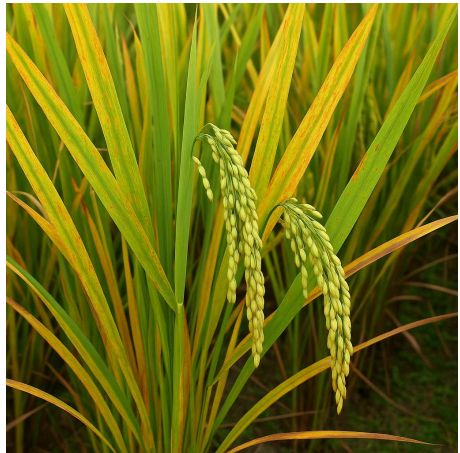 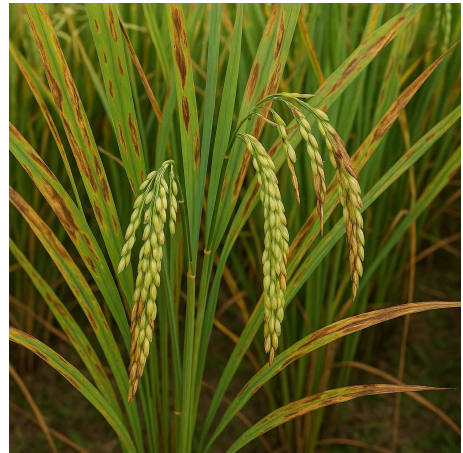

(a)      (b)

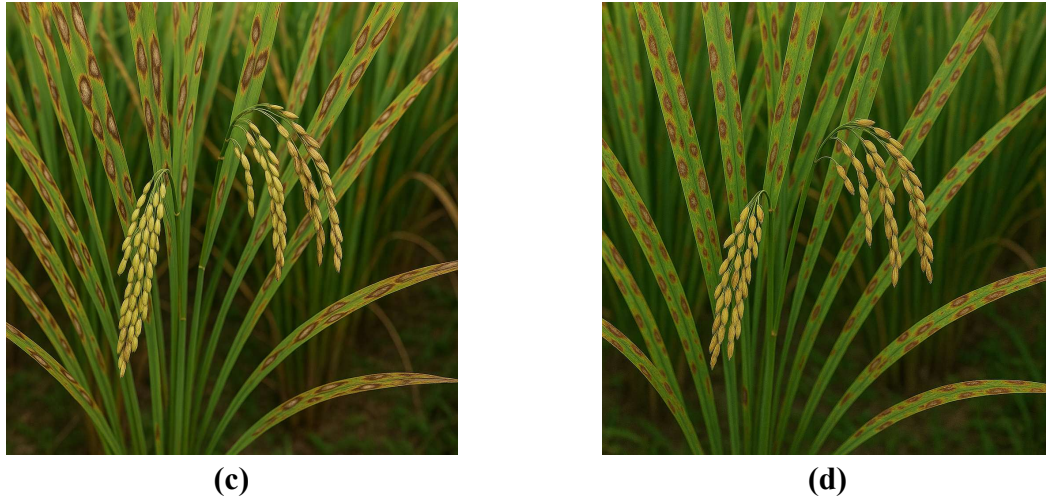

         **(c)**                 **(d)**

**Figure 1.5: Different Diseases of Rice Crop (a) Tungro Disease (b) Bacterial Blight (c) Blast Disease (d) Brown Spot Disease**

   In comparison to contemporary technological solutions, traditional manual crop disease observation is not only more time-consuming but also less accurate. The intricacy of illness symptoms and the level of skill needed for a precise diagnosis are two drawbacks of conventional approaches that may impede the development of efficient management plans. Furthermore, these expert networks might not always be easily reachable, especially in isolated agricultural regions where connectivity can be erratic. To improve agricultural resilience, this scenario emphasizes the need for more effective and trustworthy disease detection techniques. Our research focusses on using deep learning-based techniques to categorize four different disease type of diseases *(as shown in Figure 1.5)* Bacterial Blight, Brown Spot, Blast, and Tungro.

   In this work we aim to enhance the efficiency and accuracy of diagnosis of diseases using a diverse set of deep learnings. In our framework, a number of Convolutional Neural Network (CNN) architecture, including VGG19, InceptionNet, Mo bileNet, and ResNet-50, will be fused. Such architectures are adequate to identify the small differences between the healthy and diseased rice plants due to their popularity in according to robustness in identifying pictures.

Also, we are going to apply two different approaches to Explainable Artificial Intelligence (XAI), such as SHAP (SHapley Additive exPlanations) and LIME (Local Interpretable Model-agnostic Explanations) to enhance the interpretability of our deep models. Such techniques will help to explain how our models make decisions, thus allowing agronomists and farmers to understand the factors that are considered in classifying diseases. Such tools will aid in growing more confident in automated systems, as well as facilitate the process of making decisions regarding crop management through displaying the plant features which are indicative of specific disorders.

To end, our research aims to employ state-of-the-art deep learning solutions to the burning issues of crop disease identification in rice cultivation. Our goal will be to give farmers better and reliable methods of approaching the management of rice crops health through highlighting accurate classification and use of explainable AI methods. Considering the increasing population, through this combination, sustainable agricultural practices will be facilitated, and productivity will be boosted, which will eventually improve food security in the global market.

## 1.3    Significance

### 1.3.1    Significance of Rice Grain Classification:

1. **Quality Assurance:** The rice varieties require accurate classification as a point of quality benchmark to satisfy the consumers and to increase competitiveness in the market. It helps to find out premium varieties that may fetch a higher price to add value to the product.

2. **Market Differentiation:** Rice grain classification allows producers and retailers to develop their marketing strategies based on definite varieties in order to respond to the needs of the finest consumers and to develop the marketing strategies to add more to the market share.

3. **Supply Chain Efficiency:** The grading of rice grains is crucial in streamlining supply chain processes since this ensures proper deliveries to the markets. It reduces wastage and increases the efficiency of logistics within the distribution channel.

4. **Research and Development:** A better insight into the nature of the various rice types leads to research efforts aimed at breeding schemes. Such programs are intended to develop new and hardy strain that can resist various environmental problems, enhancing the development of the agricultural sector.

5. **Food Security:** Classifications can help in the process of identifying and popularizing the most appropriate type of rice to be used in different geographical locations as this can enhance better farming practices. In its turn, this enhances food security, particularly among the populations that are strongly dependent on rice as one of the primary sources of food.

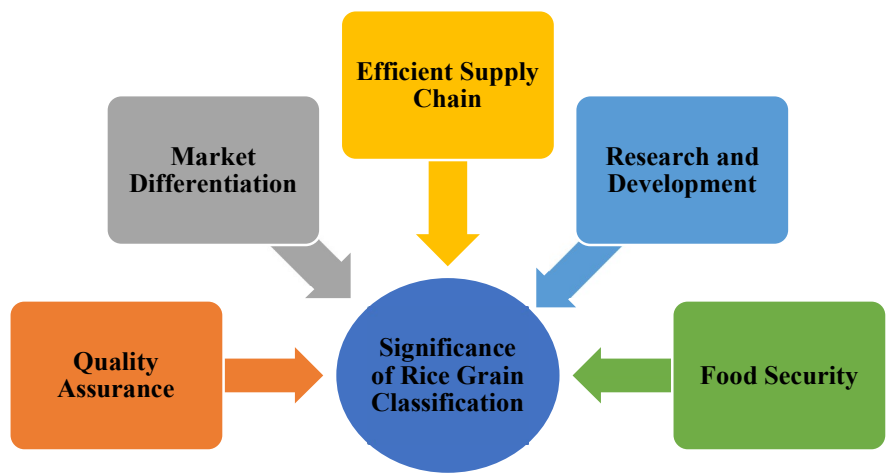

**Figure 1.6: Significance of Crowd Scene Analysis**

### 1.3.2 Significance of Rice Crop Disease Detection:

1. **Early Identification:** This is because the timely detection of diseases in rice crops can help in early interception before spread of the pathogens damages the crops and hence crop loss.

2. **Increased Yield:** Farmers are able to curb the effects of diseases in their crops through early detection thus increasing the yield and this increases profitability.

3. **Sustainable Practices:** With effective disease detection, sustainable practice in agricultural activities can occur as there is less requirement of unnecessarily applying pesticides in the farms and thereby the environment.

4. **Resource Management:** Early detection of the disease allows farmers to manage resources better like directing treatment to a few areas and also cutting their total resource expenditure.

5. **Enhanced Crop Resilience:** The knowledge about the crops disease dynamics and responses may be used to encourage crop breeding efforts that result in resilient rice crop against diseases that may be sustained over the agricultural life of rice crops.

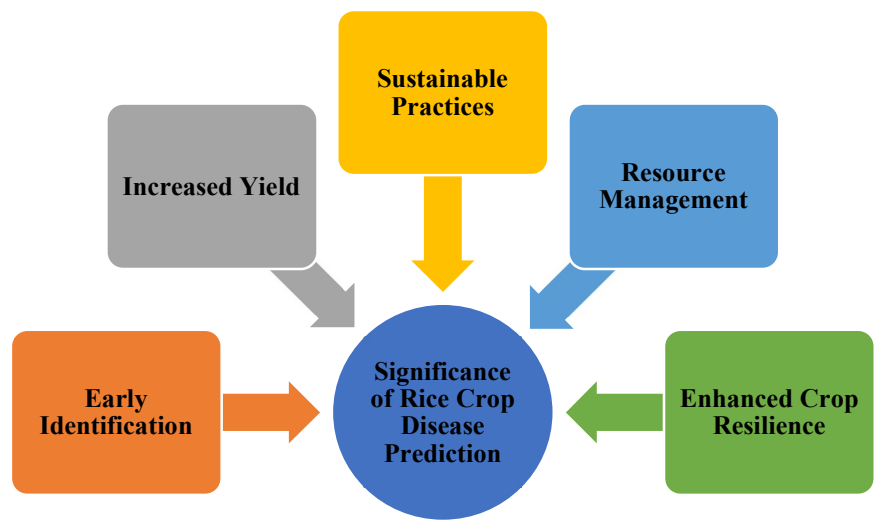

**Figure 1.7: Significance of Rice Crop Disease Detection**

## 1.4 Aims and Objectives:

The goal of the research will be to develop reliable algorithms of visual analysis of crops as well as their specific diseases. In this way, the most innovative solutions, such as computer vision and deep learning will be used, and smart solutions to detect the diseases of rice leaves and classify rice grain will be created. Although feature engineering has gotten a substantial development over a range of computer vision tasks, the aim of the given research is to explore and evaluate several features and their combinatorial approach with the focus on the crop classification and disease detection. These characteristics are considered and assessed in the study, which tries to enhance the precision and effectiveness of crop analysis algorithms.

1. The system being proposed ought to be capable of properly classifying the various types of rice grains accurately.
2. The suggested system is supposed to detect various types of diseases found in the leaves of rice crops in an accurate and early manner.

Following the objectives have been proposed to achieve this aim:

1. To propose an automatic framework based on the integration of computer vision, deep learning and explainable AI approaches for real-time crop classification.
2. To propose an automatic solution for early and accurate disease detection in rice crops.

## 1.5 Research Question:

The goals and motivations led to the following research questions

1. How can we effectively classify the diverse types of rice grains available in the market to uphold quality standards?
2. How can we analyze and detect various diseases affecting rice leaves promptly?

## 1.6 Contributions:

Particularly our contributions to crop classification and disease analysis can be summarized below:

1. The primary innovations of this study revolve around the amalgamation of explainable AI techniques, specifically SHAP and LIME, with Convolutional Neural Networks (CNN) to enhance the interpretability of deep learning models used for categorizing rice grain types as well as disease classification of rice crop. By incorporating CNNs with LIME and SHAP, our research empowers end users such as farmers, agricultural experts, and industry quality analysts to have confidence in and effectively apply these models in real-world scenarios.

2. Furthermore, our study presents a valuable contribution through the comparative assessment of LIME and SHAP. While LIME constructs localized interpretable models, SHAP offers a broader perspective on feature significance. Contrasting these methodologies sheds light on their respective advantages and limitations, delivering valuable insights for achieving optimal interpretability.

## 1.7 Stakeholders:

The following parties are involved in a rice grain classification project:

- **Farmers:** Accurate classification directly benefits farmers by enhancing crop quality control and market positioning.

- **Agricultural Researchers:** Apply knowledge to improve crop management techniques and develop agricultural technologies.

- **Food Processing Companies:** Depend on accurate classification for maintaining product quality and consistency when processing different rice varieties.

- **Retailers and Distributors:** Use accurate rice variety classification to improve marketing tactics and inventory management.

- **Regulatory Bodies:** Make use of project results to ensure adherence to food safety regulations and comply with them.

- **Consumers:** Gain from improved classification, which guarantees access to preferred rice types for gastronomic tastes.

- **Technology Developers:** Take advantage of the project's technical components to advance machine learning tools and algorithms.

- **Investors and Funders:** Seeking to provide financial support, they are interested in the project's outcomes and possible market applications.

- **Academic Institutions:** Use the project's results to conduct cooperative research and provide instruction.

Involving these parties at every stage of the project can improve its overall impact on the agriculture industry as well as its relevance and applicability.

## 1.8 Outline of the Thesis:

Based on the intended research questions, the whole work is divided into seven chapters. The thesis is structured as below:

- Chapter one begins by introducing the crop analysis, rice crop classification, crop disease detection, and explaining its importance in different applications of the real-time field. It provides a detailed introductory overview of rice grain classification and rice crop disease detection. It outlines the motivation behind research carried out in the field of crowd scene analysis. It addresses the major challenges that must be overcome to meet the research goals. It concludes by summarizing the research contributions.

- Chapter two highlights the comprehensive review of available research on the research topic. It investigates prior studies, theories, models, and methodologies that have been already proposed in the subjected discipline. This chapter also establishes groundwork by highlighting research gaps in current knowledge, and by emphasizing key findings and comments from prior publications.

- Chapter three highlights the proposed framework for rice grain classification using xAI techniques. This chapter is divided into three parts. The first part highlights the theoretical background of rice grain classification and outlines a detailed overview of deep learning, neural networks, data augmentation, convolutional neural networks, and explainable AI techniques. The second part of this chapter describes the end-to-end proposed framework. It explains the experimental design, data collection, and pre-processing techniques. The part outlines the specific algorithms, or techniques that will be used to address the research questions.

- Chapter four presents and discusses the implementation as well as experimental results for rice grain classification. The first part of this chapter highlights the implementation details, evaluation metrics, and data sets. However, the second part includes experimental results, performance evaluations, comparisons with existing methods, and any statistical analyses conducted. The chapter interprets the results, draws conclusions, and discusses the implications of the research questions. It also highlights any limitations or challenges encountered during the experiments.

- Chapter five highlights the proposed framework for rice crop disease detection using xAI techniques. This chapter is also divided into three parts. The first part highlights the theoretical background of crop disease detection and outlines the VGG-19, RESNET-50 architecture of CNN and their integration with xAI models such as LIME and SHAP. The second part of this chapter describes the end-to-end proposed framework. It explains the experimental design, data collection, and pre-processing techniques. The part outlines the specific algorithms, or techniques that will be used to address the research questions. Implementation details, evaluation metrics, and data sets are presented in the third part of this chapter.
- Chapter six presents and discusses the implementation as well as experimental results for rice crop disease detection. The first part of this chapter highlights the implementation details, evaluation metrics, and data sets. It includes performance evaluations, comparisons with existing methods, and any statistical analyses conducted.
- Chapter seven presents the conclusions and future work by summarizing the main findings and contributions of the research.

# CHAPTER TWO: LITERATURE REVIEW

A brief review of the current approaches along with their shortcomings for crowd scene analysis is presented in this chapter. The background review is divided into two parts, with each section focused on crop classification and crop disease detection. This detailed discussion will highlight the limitations and research gaps to find research direction for future work.

## 2.1    Crop Classification:

With the recent advancements in the field of artificial intelligence, deep learning approaches are utilized in various industries, from airlines to predict the behavior of passengers [45] to the predictive maintenance of machines in industries [46]. It has also been utilized in different classification tasks such as cancer cell classification [47], [48], [49], [50], patient classification [51], [52], [53].

Deep Learning approaches also play a significant role in agriculture [29]. There is a lot of research carried out in the domain of crop analysis by leveraging the use of deep learning and machine learning techniques with a special focus on crop classification [54], [55], [56] and crop disease prediction [56], [57], [58], [59]. Rice grain classification remains a significant challenge for farmers and quality analysts despite the existence of various proposed and proven models. This research project presents an automatic framework for efficient and accurate classification of rice varieties, including arborio, jasmine, basmati, ipsala, and karacadag, based on their morphological features. This section explores the literature on deep learning approaches for classifying rice grains. In addition, the features of our proposed model are discussed at the end of this section

Koklu et al. [60] compared the performance of ANN and DNN for the feature-based datasets and CNN for the image-based datasets. Models were evaluated based on seven metrics such as F1-Score, Precision, accuracy, FPR, FNR, sensitivity and specificity. There was a dataset (image-based) with an amount of 75K images of five different types of grains of rice including arborio rice, jasmine rice, basmati rice, ipsala rice, and karacadag rice that was used to train the models. Average classification accuracy of grain was shown in the study to be 100 percent, 99.95 percent, and 99.87 percent, respectively, in the case of CNN, DNN, and ANN, and this research could also classify various forms of food grains.

In [61], An author has carried out a study on the different machine learning classifiers whose results are logistic regression, decision tree, support vector machine, random forest, multi-layer perceptron (MLP), Naive Bayes (NB), and KNN on the classification of different varieties of rice. To test the model, the author used performance indicators in the form of F1-score, accuracy, precision and recall. The performances of the experiments were encouraging whereby random forest classification had 99.85 percent accuracy and decision tree classification had 99.68 percent accuracy.

In [62], A different author presented a framework, depending on a multi-class SVM, of classification of three types of rice grains namely basmati, ponni, and brown grains. The shape and color of grains allowed extracting four various features, including length of major axis, length of minor axis, area, and perimeter using the regionprop() function. The proposed study was tested on 90 testing images and obtained 92.22% of accuracy and classification.

Ramadhani et al. [63] offered an article to maximize the efficiency of SVM and KNN through the use of genetic algorithms. They had a 92.81 percent accuracy of the SVM-GA and 88.31 percent accuracy of the KNN-GA. The findings have demonstrated that SVM combined with Genetic Algorithms were better than anything and may enhance sustainable agricultural practices in rice growing and contribute to raising the general productivity [64].

A combination of Deep Convolutional Neural Network (DCNN) with Support Vector Machine (SVM) offered by Bejerano et al [65]. The training framework suggested in this paper was used to classify four types of rice, namely damaged rice grain, discoloured rice grain, broken rice grain and chalky rice grain. The proposed classification was through SVM, whereas morphological features were extracted using DCNN. The model was able to rate the rice grading and classify it with a 98.33 percent training classification and 98.75 percent validation rates.

Rayudu et. al. [66] suggested a framework in the form of RESNET-50 to classify five types of rice grains such as basmati, jasmine, arborio, ipsala and karacadag. The testing and training process of the model was done on a small database with 2500 images (500 images per each of the classes). The data was isolated into 80:20 ratio and the results subsequently compared with those of VGG16 and MobileNet. The proposed comparative study revealed that the RESNET-50 performed best when it came to the process of rice categorization, which indicated the importance of deep learning in enhancing the process of rice categorization to bring out its categorization efficiency thus increasing its level of quality control, supply chain management, and consumer confidence.

In [67], method using a combination of Convolutional Neural Networks (CNNs) and transfer learning was suggested to classify rice types and identify them properly. The suggested model relies on the multiple architectures of CNN such as MobileNet, ResNet50, and VGG16. Architectures were employed in the transfer learning approach in improving the general performance of the model. The transfer learning with the MobileNet in the CNN model delivered an excellent accuracy score of 98.94%, whereas ResNet50 and VGG16 registered 79.79% and 99.47% accuracy rate, respectively. The high rates of accuracy point out the prospects of the model in real-time application of rice variety classification and first-class evaluation.

In [68], another author proposed a framework based on a combination of CNN with transfer learning of MobileNetV2 for the classification of rice grain. Results revealed that the MobileNetV2 model outperforms the other model due to its superior feature extraction from large-scale ImageNet dataset images.

In [69], an author presented a framework to assess machine vision methods for the classification of six Asian rice varieties: Super-Basmati-Kachi, Super-Basmati- Pakki, Super-Maryam-Kainat, Kachi-Kainat, Kachi-Toota, and Kainat-Pakki. The dataset was collected by gathering 10,800 rice grain samples from Bangladesh, India, China, Pakistan, and other nearby nations. The features of each image were retrieved after converting the photos to an 8-bit grayscale format. LMT-Tree had the highest overall accuracy (MOA) of all five machine vision classifiers, at 97.4%. 97.4%, 97.0%, 96.3%, 95.74%, and 95.2% were the classification accuracies of LMT Tree (LMT-T), Meta Classifier via Regression (MCR), Meta Bagging (MB), Tree J48 (T-J48), and Meta Attribute Select Classifier (MAS-C).

Table 2.1: Summary of Recent Studies on Rice Grain Classification Using Machine Learning and Deep Learning Techniques

| Ref | Model(s) Used | Rice Varieties | Features/Techniques | Performance Metrics | Best Result(s) |
| --- | --- | --- | --- | --- | --- |
| [60] | ANN, DNN (feature-based), CNN (image-based) | Arborio, Jasmine, Basmati, Ipsala, Karacadag | Feature and image-based datasets (75K images) | F1, Precision, Accuracy, FPR, FNR, Sensitivity, Specificity | CNN: 100%, DNN: 99.95%, ANN: 99.87% |
| [61] | LR, DT, SVM, RF, MLP, NB, KNN | Not specified | Machine learning classification | Accuracy, Precision, Recall, F1 | RF: 99.85%, DT: 99.68% |
| [62] | Multi-class SVM | Basmati, Ponni, Brown | Features: Major/minor axis, area, perimeter (regionprops) | Accuracy | 92.22% (90 test images) |
| [63] | SVM-GA, KNN-GA | Not specified | Optimization using Genetic Algorithms | Accuracy | SVM-GA: 92.81%, KNN-GA: 88.31% |
| [65] | DCNN + SVM (Hybrid) | Damaged, Discolored, Broken, Chalky | Morphological feature extraction via DCNN, SVM for classification | Training/Validation Accuracy | Train: 98.33%, Val: 98.75% |
| [66] | ResNet-50, VGG16, MobileNet | Basmati, Jasmine, Arborio, Ipsala, Karacadag | Deep learning comparison (2500 images, 80:20 split) | Accuracy | ResNet-50 outperformed others |
| [67] | CNN (MobileNet, ResNet50, VGG16) + Transfer Learning | Not specified | CNN with Transfer Learning | Accuracy | VGG16: 99.47%, MobileNet: 98.94%, ResNet50: 79.79% |
| [68] | CNN + MobileNetV2 (Transfer Learning) | Not specified | Transfer learning using MobileNetV2 on ImageNet features | Accuracy | MobileNetV2 outperformed others |
| [69] | LMT-T, MCR, MB, T-J48, MAS-C | Super-Basmati-Kachi, Super-Basmati-Pakki, Super-Maryam-Kainat, Kachi-Kainat, Kachi-Toota, Kainat-Pakki | Grayscale image conversion, machine vision classifiers | Accuracy | LMT-T: 97.4%, MCR: 97.0%, MB: 96.3%, T-J48: 95.74%, MAS-C: 95.2% |

## 2.2 Crop Disease Classification:

The global population is increasing, and half of the world's population relies on paddy for food. Early detection of rice leaf diseases like blast, bacterial blight, brown spots and tungro can be achieved using transfer learning models. A Comparative study was proposed by Ahmad et.al. [70] to identify three common diseases: leaf smut, bacterial leaf blight, and brown spot by leveraging different machine learning techniques such as KNN, J48, NB, and LR.

Proposed models were trained on affected leaves images with white background. Results showed that the decision tree algorithm achieved an accuracy of over 97% after 10-fold cross validation after 10-fold cross validation on the test dataset. In [56], an author proposed DenseNet201, a transfer learning technique, to identify these diseases in rice crops. The study uses 240 images and achieves a higher accuracy of 96.09% compared to existing models, and 62.20% accuracy for simple CNN. A custom memory efficient Convolutional Neural Network (CNN) called RiceNet [71] was presented, which can automatically detect rice grain diseases. Proposed model achieves a classification accuracy of 93.75%. The researchers also apply transfer learning to achieve 97.94% accuracy through EfficientNetB0, demonstrating the effectiveness of the proposed custom network in detecting and classifying diseases in agricultural crops.

A method for classifying rice leaf diseases using deep learning approaches was proposed by [72]. Machine and ensemble learning classifiers were used, but compared to CNN and transfer learning models, InceptionResNetV2 showed the highest validation accuracy of 88%. Transfer learning models outperform machine learning classifiers, highlighting the importance of continuous plant monitoring and early disease detection in agriculture.

In [73], a CNN-based method for identifying and categorizing three prevalent rice leaf diseases—Bacterial Leaf Blight, Brown Spot, and Leaf Smut—was developed. The suggested CNN model outperformed other variations with an accuracy of 87% and a loss of 0.75, even without dropout and batch normalization. A method based on combination of CNN and SVM was proposed by Rajmohan et. al. [74]. Different lesion shapes and colors are essential for identifying paddy leaf disease, which is divided into Nematode, Blast, Smut, and Spots.

These features are extracted by leveraging the use of CNN and SVM for classification of different diseases. Results showed that the proposed model achieved an overall accuracy of 87.5%. Another research based on combination of CNN and SVM was proposed by Liang et. al. [75]. Results showed that the proposed model showed higher level accuracy results. The proposed study also showed that CNN extracts high-level features more effectively and discriminatively than Haar-WT (Wavelet Transform) and local binary patterns histograms (LBPH). The quantitative evaluation also shows that CNN with SoftMax and CNN with support vector machine perform similarly, with better ROC curves, larger AUC, and higher accuracy than LBPH+SVM and Haar-WT+SVM.

Okten et. al [76] proposed a computer-based approach that was meant to recognize and decide diseases in rice plants. The system will have four steps including pre-processing of images, segmentation, feature extraction and classification. The pre-processing is conducted using the median filtering method, and afterward, it is implemented OTSU to segment the images. Features are extracted using GLCM method and the possibility of whether the plant is diseased is determined using Probabilistic Neural Network (PNN). The accuracy of the system that is proposed is 76.8%.

In [73], deep learning as a tool of early detection and classification of wheat diseases on two datasets, Global Wheat Head Detection (GWHD) dataset to detect the object and Large Wheat Disease Classification Dataset (LWDC) as the dataset to classify the object. On LWDC, a YOLOv4 object detection network trained on GWHD attains 91 percent mean Average Precision (mAP), and the network performs domain transfer. Five CNN models, such as VGG19, are compared in the task of the classification of wheat diseases, and the best-performing model is VGG19.

Nevertheless, despite the high overall accuracy during the classification, the methods, such as CNNs and preprocessing methods, might not be effective to explain their predictions well enough. Such uninterpretability may be a severe disadvantage, particularly in such areas as agriculture where the stakeholders should have the rationale of the model outputs in order to make an informed decision. Users might not be able to trust and successfully use the technology in impractical scenarios without being clearly explained why a specific classification was made, including what characteristics of the grains of the rice or plants of the wheat the model relied on. Consequently, it is highly important that explainability methods be included to narrow down the said gap, which includes LIME and SHAP. These techniques have the ability of explaining the process of decision making in terms of identifying the particular attributes that influenced the model to provide such ratifications and so making the users more confident as well as adopting better management solutions to dealing with the rice grain species and the disease diagnosis in wheat plants. Interpretability of our models could help us to make sure that our models are not only correct but also viable and credible in the context of real-life agricultural uses.

Table 2.2: Comparative Review of Machine and Deep Learning Approaches for Rice Leaf Disease Detection and Classification

| Ref | Model(s) Used | Diseases/Target | Techniques/Features | Dataset Info | Performance |
|---|---|---|---|---|---|
| [70] | KNN, J48, NB, LR | Leaf Smut, Bacterial Blight, Brown Spot | ML models trained on leaf images with white background | Not specified | J48: >97% (10-fold CV) |
| [56] | DenseNet201, CNN | Rice Leaf Diseases | Transfer learning vs. simple CNN | 240 images | DenseNet201: 96.09%, CNN: 62.20% |
| [71] | RiceNet (custom CNN), EfficientNetB0 (transfer learning) | Rice Grain Diseases | Custom lightweight CNN with transfer learning | Not specified | RiceNet: 93.75%, EfficientNetB0: 97.94% |
| [72] | CNN, InceptionResNetV2, ML & Ensemble classifiers | Rice Leaf Diseases | Deep learning vs. machine & ensemble models | Not specified | InceptionResNetV2: 88% validation accuracy |
| [73] | CNN | Bacterial Blight, Brown Spot, Leaf Smut | CNN without dropout/batch normalization | Not specified | Accuracy: 87%, Loss: 0.75 |
| [74] | CNN + SVM | Nematode, Blast, Smut, Spots | Lesion shapes and color features | Not specified | Accuracy: 87.5% |
| [75] | CNN + Softmax, CNN + SVM, Haar-WT + SVM, LBPH + SVM | Rice Leaf Diseases | CNN vs. traditional methods for feature extraction and classification | Not specified | CNN-based models outperformed traditional methods |
| [76] | PNN (Probabilistic Neural Network) | General Rice Plant Diseases | 4 stages: Median filter, OTSU, GLCM feature extraction, PNN classifier | Not specified | Accuracy: 76.8% |
| [73]* | YOLOv4 (object detection), VGG19 (classification) | Wheat Head and Disease Detection | Domain transfer learning from GWHD to LWDC datasets | GWHD, LWDC | YOLOv4 mAP: 91%, VGG19 top classifier |

# CHAPTER THREE: CROP CLASSIFICATION METHODOLOGY

## 3.1 Theoretical Background

### 3.1.1 Introduction to Deep Learning

Artificial intelligence (AI) is a subfield of computer science, which involves the development of the machines that are capable of doing the tasks that usually need human intelligence. It has become very necessary in many practical applications such as robotics, computer vision, machine learning and natural language processing. The term AI is a general term embracing a range of techniques and means to assess vast amounts of data, distinguish their trends and enhance the capabilities to make predictions with ever more high-quality and accuracy.

Machine learning is a subset of artificial intelligence that investigates the computer algorithms and models that enable the machine to combine analytical, learning, and decision making by selecting the relevant features manually input data. The selection of relevant characteristics is always applied to influence the performance level of a model positively. Labeled detests train models.

Deep learning as a subset of machine learning tends to involve more powerful and complex algorithms having many hidden layers. Algorithms that learn deep learning are able to identify complex patterns and connections found in data by way of automatic learning and building hierarchical depictions of the data. Deep learning is thus particularly useful in such realms as computer vision and natural language processing. Deep learning involves neural networks, in which learning and making predictions takes place. Deep learning is a data-hungry paradigm that always requires significant amounts of data as compared to machine learning.

This ability can automatically discover any relevant patterns and create accurate predictions with the help of deep learning models' assistance [107].

### 3.1.2 Neural Networks

Deep learning methods are founded on the neural networks which constitute a narrower branch of machine learning Artificial NN (ANNs) and Simulated NN (SNNs). These networks were inspired by the way the human brain was created and how it functions, as this organ is composed of interlinked networks of small neurons that transfer signals. Neural networks **(as shown in** Figure 3.1: Architecture of a Simple Neural Network*Figure 3.1*) are adaptive, therefore machines can correct themselves based on the mistakes and enhance their performance. Neural networks enable machines to analyze data, discover patterns and adjust their own algorithms based on the information used. Leveraging on this iterative process of learning enables machines to solve complex problems as well as more accurately predict outcomes among other things, making them gradually augment their output [77], [78].

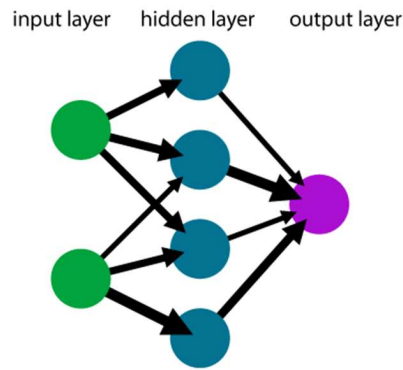

**Figure 3.1: Architecture of a Simple Neural Network**

Computer models based on the human brain are known as artificial neural networks (ANNs) *(as shown in Figure 3.2)* , those consist of a number of layers of nodes that are interconnected inseries, and such structures are also known as artificial neurons. These networks are composed of three main types of layers namely the input layer, the hidden layer and the output layer. The importance of one layer to the functioning of the network and its effectiveness is very vital [79].

### 5.1.1 Structure of Artificial Neural Networks

1. **Input Layer:** That is the input layer of the ANN where data is given as input. Each node on this layer represents a characteristic or a feature of input data. The nodes can correspond to pixel value in image recognition problems [80], [81].

2. **Hidden Layers:** There is input layer, output layer, and one or more hidden layers in between these. These layers are important in the capability of the network to identify complex patterns. Each of the hidden layers processes the inputs of the preceding layer by a number of nodes. Complexity of the task may influence the number of hidden layers and nodes that shall be utilized. The network can learn higher-order abstractions and networks in the data when there are more hidden layers [80], [81].

3. **Output Layer:** The last layer of calculation is the output layer that produces the result of the network. From this layer each node reflects a different class or the projection of the result. As an example, one or two output nodes can correspond to the two classes in a two-class task of a classification [80], [81].

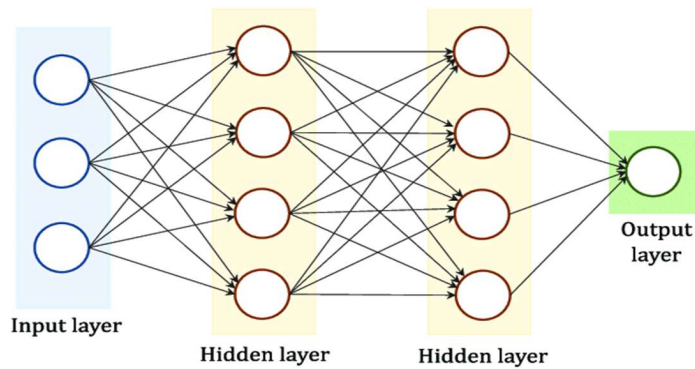

**Figure 3.2: General Architecture of Artificial Neural Network**

### 5.1.2 Functioning of Artificial Neural Networks

Artificial neural network (ANN) has nodes through which calculated mathematical functions referred to as activation functions are used to process the input data. These functions calculate such output of a node based on the weighted sum of the inputs of a node [80], [81]. Common activation functions consist of:

- **Sigmoid:** It is widely applied in the binary classification tasks and input values are mapped by a range of 0 to 1 [82], [83], [84].
- **ReLU:** The Rectified Linear Unit (ReLU) is rectified, that is, it outputs the input in the case there is a positive input or zero in other cases. This in the deeper networks reduces the outcomes of the vanishing gradient problem [84].
- **SoftMax:** This is a common method used in cases of multi-class classification which converts raw values outputting into probability [85].

### 5.1.3 Weights and Learning Process

Artificial neural network learning weights assigned to node to node connection matter as they show the strength of connection. These weights are constantly adjusted to enhance the network performance regarding accuracy and predictive powers. This is a learning process, which passes through several basic stages:

- **Forward Propagation:** Layer by layer, input data moves across the network. Using its activation function, each node processes inputs before sending the output to the layer below it, until it reaches the output layer.
- **Loss Calculation:** After output is generated, the network uses a particular loss function to compare the expected output with the actual target values in order to determine the loss or error. This function measures how well the network's predictions match the intended results.

- **Backward Propagation:** The advantages of backpropagation (as shown in **Figure 3.***3*) [14] is that it enhances accuracy by computing the gradient of the loss field concerning the weight that determines the network. The gradients indicate the contribution that the weight plays to the overall error.

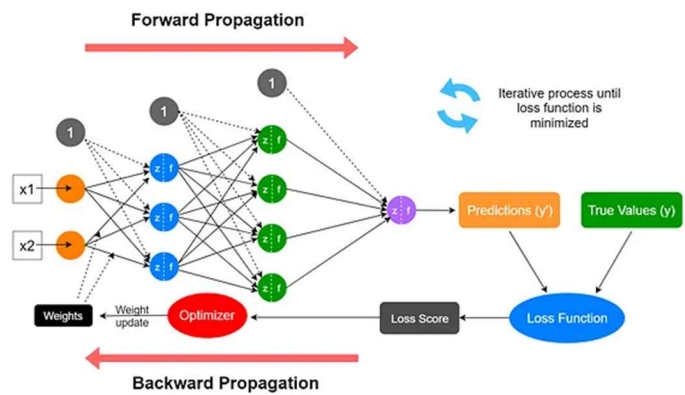

**Figure 3.3: Backpropagated Neural Networks – How it works?**

- **Weight Update:** The updating of weights is done opposite to the gradient with the optimization algorithms such as Adam or Stochastic Gradient Descent (SGD). This modification is aimed at minimizing future predictions losses. This iterative process goes on until the model attains a pre-set level of training iterations or converges to the desired set of weights.

To sum up, because of layered network structure and the learning mechanism, artificial neural networks are exceptionally efficient tools that allow resolving classification and prediction problems due to their ability to comprehend the interdependency of complex data. ANNs enhance their capacity to make correct predictions in that weights are modified often in light of input information as well as feedback. This reasons why they are priceless in many areas ranging to the image recognition and the natural language processing, among others. Capabilities and structures of ANNs evolve as a result of research and technology effectively increasing the real world applicability to solving real world problems.

### 3.1.3 Convolutional Neural Networks

The primary objective of artificial intelligence is to narrow the disparity between human and machine capabilities [14]. Deep neural networks excel in recognizing complex patterns and relationships, contributing to achieving this goal. Furthermore, the deep hidden layers of Deep NNs have surpassed the performance of traditional machine learning techniques [14]. Convolutional neural networks, known as ConvNets, are a specific type of deep neural network extensively utilized in computer vision for tasks involving the analysis, learning, and prediction from visual data like images and videos [86]. General architecture of CNN is *(as shown in Figure 3.4)*.

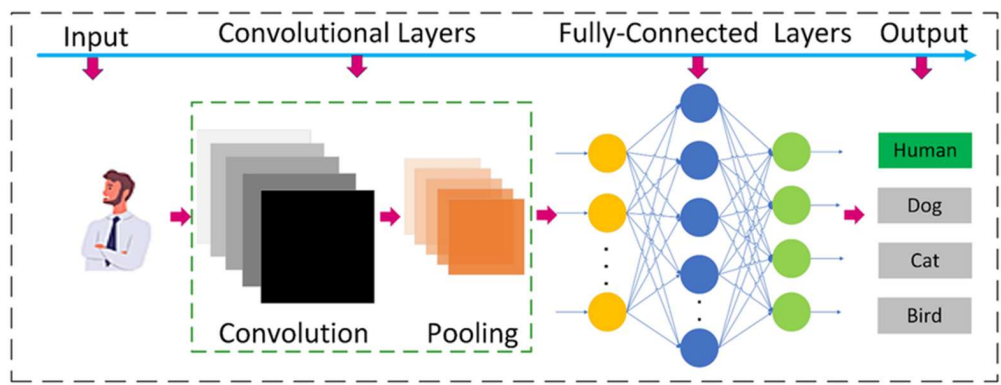

**Figure 3.4: General Architecture of Convolutional Neural Network**

- **Input:** The initial stage of a convolutional neural network (CNN) involves taking images as input, processing them through hidden layers, assigning weights and biases to crucial features, and preparing them for prediction or classification tasks. The CNN typically consists of three main segments: input, feature extraction, and output [86], [87].

- **Image Input:** The input phase receives images and directs them to the feature extraction network. The number of nodes in the input layer matches the pixels in an image, with each pixel representing an individual input to the network [86], [87].
- **Feature Extraction:** This network usually comprises convolution, activation, and pooling layers, with varying numbers of these layers affecting the network's depth and performance. More layers lead to enhanced network performance due to increased depth [86], [87].
- **Convolution Layers:** These layers execute the convolution operation by applying a filter to the image to extract essential information, producing feature maps displaying specific features at different spatial locations [86], [87].
- **Pooling Layers:** Feature maps undergo processing in pooling layers to decrease their size, reducing complexity and computational requirements. Pooling methods include max pooling (selecting maximum values within a window) and average pooling (calculating the average of values within the window) [86], [87].

The evolution of Convolutional Neural Network (CNN) architecture has been transformative in the realm of deep learning, leading to significant progress and advancements in various applications over the years.

The progression of CNN Architectures includes:

1. **Traditional CNNs:** LeNet-5 started it all with basic layers for sorting images. It showed us how deep learning could help computers understand pictures.

2. **AlexNet:** came out in 2012. It went deeper than before, used ReLU to make things work better, and added dropout to stop overfitting. It did well with ImageNet pictures and got everyone excited about CNNs [88], [89].

3. **VGGNet:** VGGs kept things simple but effective. It used lots of small 3x3 filters stacked on top of each other. It worked great and wasn't too hard to use [88].

4. **Inception Networks:** GoogleNet tried something new with "inception modules." These let the network look at pictures in different ways at the same time. It worked better without needing tons more computer power [46].

5. **ResNet:** It added shortcuts between layers so information could flow better. This meant we could make deep networks - hundreds of layers - without the usual training problems [88], [90].

6. **MobileNet:** These types of CNNs work better on phones and small devices. It changed how the math worked to use less power and memory but still get good results [91].

7. **EfficientNet:** These types of networks came up with a smart way to make networks bigger. It carefully grew the network's depth, width, and image size together. This meant better results with fewer numbers to crunch.

These improvements in CNN design have changed everything about how computers see things. They've made it much easier for machines to sort pictures, find objects, and watch videos. Now we can train bigger and more complex systems using huge amounts of data, which helps them work better at all sorts of different jobs. The changes haven't just made things a bit better - they've completely changed what's possible. Thanks to these better systems, computers are way more accurate at understanding what they're looking at, whether they're picking out faces in a crowd or spotting problems in medical scans. It's amazing how far we've come with these tools.

## 3.2 End-to-End Proposed Framework for Rice Grain Classification:

The proposed approach aims at improving the identification of different varieties of rice grains using the Convolutional Neural Networks (CNN) and explainable AI tools, specifically Local Interpretable Model-agnostic Explanations (LIME) and Shapley Additive Explanations (SHAP). Data acquisition, data division in terms of training, validation, testing, exploratory data analysis, pre-processing activities all form part of the system. This is followed by subjecting the resulting data to CNN training, after which it automatically identifies relevant features in the raw data and presents the data in a hierarchical form. Then two different explainable AI methods can be applied to the labeled pictures: SHAP which delivers the explanation on the global-level perspective of CNN performance and LIME which delivers the explanation on the local-level perspective of the CNN performance. The performance of the final model will be displayed by using the predetermined metrics on a test set. Besides improving classification accuracy, this combination of phenomena facilitates transparency and trust in decision-making. The proposed system aims at achieving accuracy in classification of rice grains. The proposed system for rice grain classification is *(as shown in Figure 3.5).*

## 3.3 Description of Proposed System:

### 3.3.1 Image Acquisition:

For system development, the Rice Image Dataset [47], which is openly accessible on Kaggle, is suggested. There are 75,000 photos in the suggested dataset, with 15,000 photos for every type of rice grain.

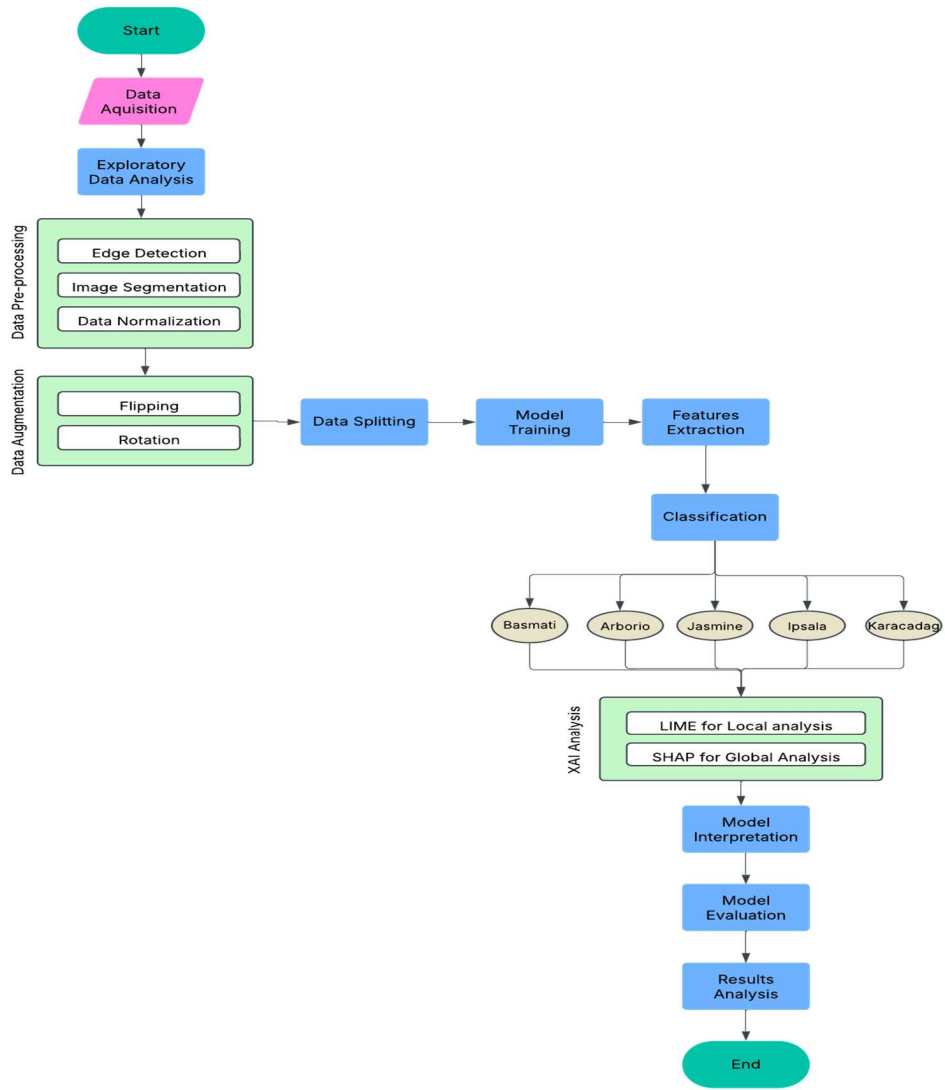

**Figure 3.5: Proposed architecture for classification of different varieties of rice grains.**

### 3.3.2   Exploratory Data Analysis (EDA):

Exploratory Data Analysis (EDA), a visual technique for analyzing data distributions, relationships, and anomalies, is applied to the obtained images. More dependable analytical results are produced by this process, which also helps identify potential problems like missing values and outliers and informs feature selection, data pre-processing, and model development.

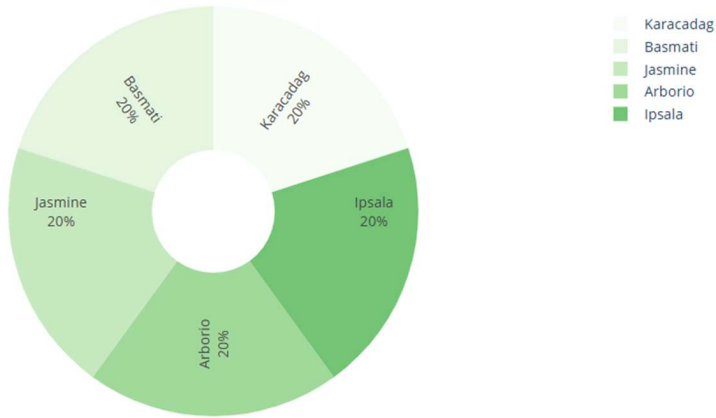

**Figure 3.6: Class Distribution of Different Varieties of Rice Grains**

Training, testing, and validation comprise the three sections of the dataset. The remaining 20% of the data is split equally between testing and validation, with the remaining 80% going toward training. Following splitting, the data distribution will be *(as shown in Figure 3.7).*

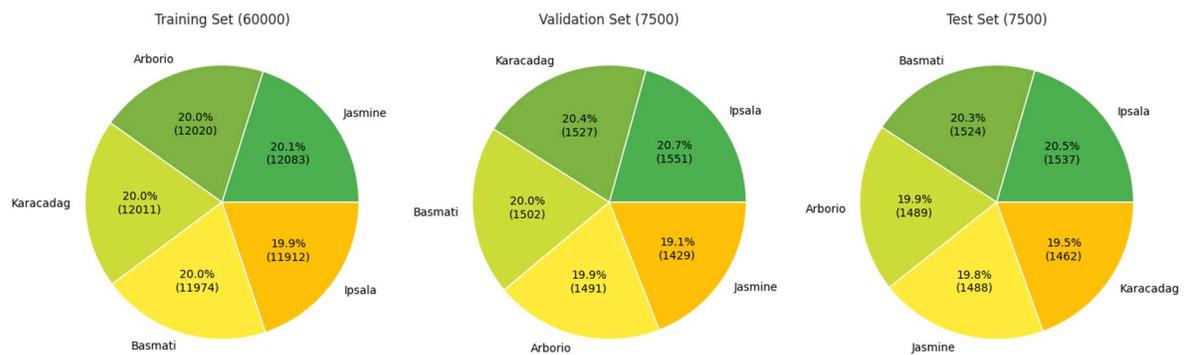

**Figure 3.7: Dataset Distribution after Splitting**

### 3.3.3 Image Preprocessing:

#### *3.3.3.1 Edge Detection:*

To efficiently identify grain boundaries in images *(as shown in Figure 3.8),* edge detection will be applied, making further classification and analysis easier. In this study, we used Canny edge detection, a reliable method that is especially well-suited for defining the contours of rice grains because it can detect abrupt changes in image intensity.

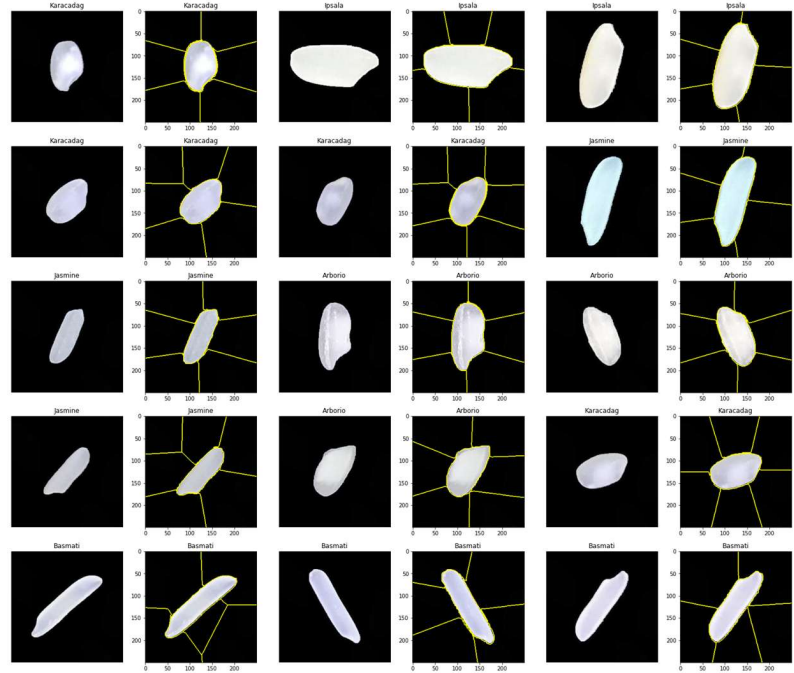

**Figure 3.8: Rice grains with detected boundaries**

The existence of lines around rice grains after Canny edge detection can be explained by low threshold configurations, noise, and faint gradients that cause extraneous edges to appear. By using Gaussian blurring to reduce noise, non-maximum suppression to accurately outline edges, and hysteresis thresholding to connect weak and strong edges, canny edge detection seeks to preserve uniformity in size and shape. Together, these techniques produce a more accurate outline of the grain boundaries, which is essential for effective classification in later analytical procedures.

To improve edge detection accuracy and lower noise, the original images were first converted to grayscale and then subjected to Gaussian blurring. We were able to highlight the unique shapes of the rice grains by successfully extracting the edges using the Canny algorithm *(as shown in Figure 3.9)*. This technique enhances the grains' visual representation and is a crucial pre-processing step that raises the classification accuracy of rice varieties.

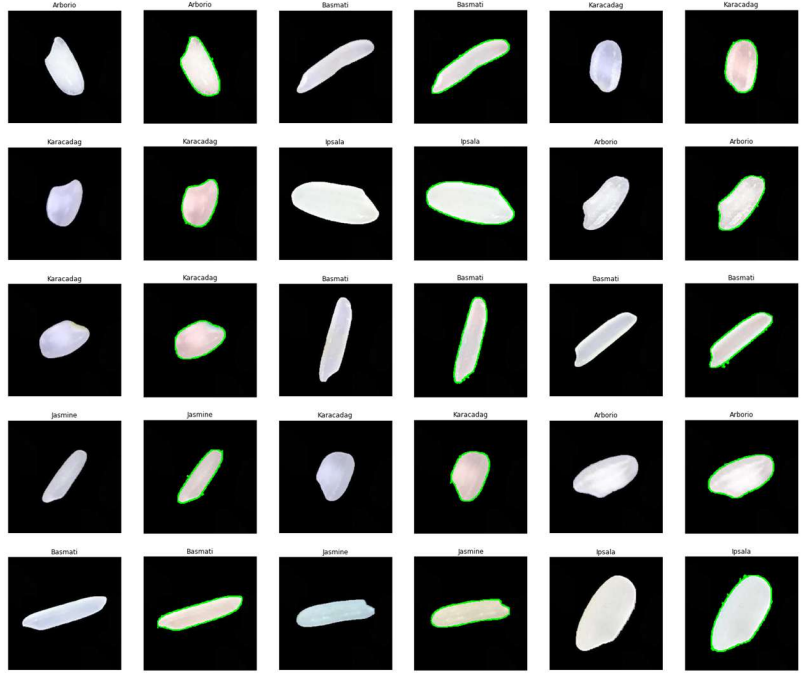

**Figure 3.9: Rice grains with detected edges, highlighting the contours and characteristics essential for classification and analysis**

### 3.3.3.2  *Image Segmentation:*

After the edge detection procedure, the images will be subjected to image segmentation to help identify various segments in an image and help compare and classify rice grains better. Image segmentation can be defined as the process, through which one can divide an image into significant parts that make sense and contain different objects or regions of interest. Through this procedure our ability to distinguish rice grains in relation to one another and the background is significantly increased resulting in the individuation of rice grains and their characteristics being raised. By splitting the grains using a siphon we will be able to enhance the general success of the ensuing classification activities by ensuring better and more reliable scrutiny of the rice varieties. Canny Edge Detection is employed in order to ensure uniformity on the size and pattern of each rice grain. This complete procedure enhances our understanding of the morphological characteristics of all rice types besides helping with proper classification.

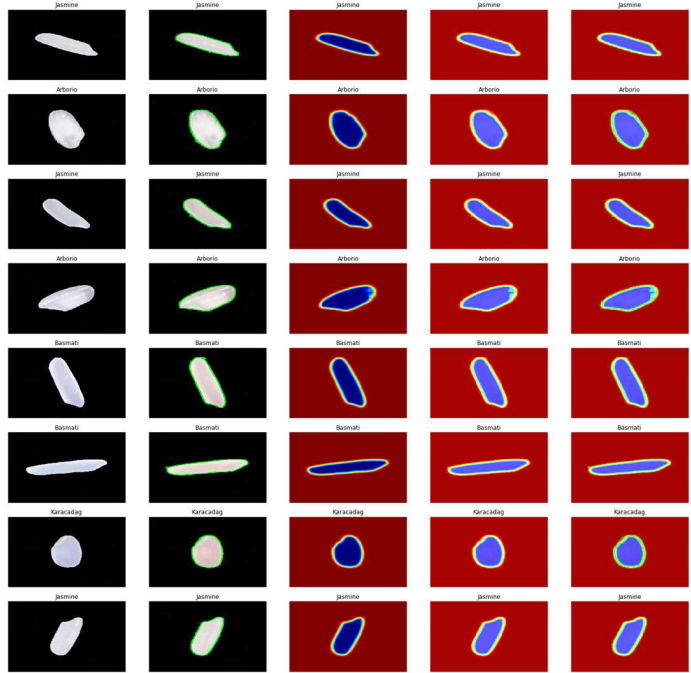

**Figure 3.10: Segmented images of rice grains for enhanced analysis**

### 3.3.3.3 *Data Normalization:*

The model struggles to converge, so during training it is very difficult due to varying pixel intensity. To address this problem all images that have been segmented will be piped into the normalization process. One of the very important steps is to scale the pixel values of all the images into range 0-1. Also, this uniform scaling enhances stability and performance in training process to characterize rice varieties more accurately and contribute to powerful findings in agricultural image analysis.

### 3.3.3.4 *Data Augmentation:*

To improve the volume and diversity of the data, two distinct data augmentation techniques — rotation and flipping—were used on every image. Images were flipped at a rate of -1 to +1 and rotated at 90 and 180 degrees.

### 3.3.4 Proposed Model:

The convolutional neural network (CNN) *(as shown in Figure 3.11)* presented in this research is meticulously de-signed for image classification, specifically targeting images of size 50x50 pixels with three channels (RGB). The architecture is constructed using the Keras Sequential API, which simplifies the process of stacking layers linearly, allowing for a clear and concise representation of the model's structure.

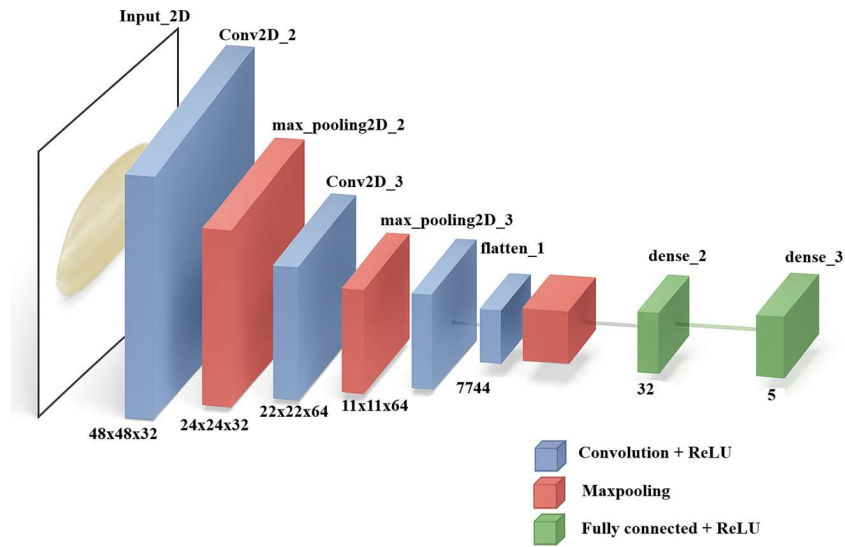

Figure 3.11: Architecture of Proposed Model for Rice Grain Classification

The model begins with the convolutional layer (Conv2D), which is learning such spatial features as edges and textures with the help of 32 3x3 filters. Dense patterns can be recognised with the help of non-linearity of ReLU activation function. It is then followed by a max pooling layer (MaxPooling2D), which enhances the balance between computing speed and generalization pattern by selecting the largest value in a 2x2 area, therefore, decreasing the features map sizes. In order to obtain representations at higher levels, another convolutional layer (64 filters) is added to the architecture, preceded again by a max pooling layer.

The flattening layer transforms the multi-dimensional output into one-dimensionalVector following pooling and provides input to the first fully connected layer that transforms learned features to abstract representations in the form of 32 units with ReLU activation. The last dense layer is made of five units, all of which have a SoftMax as an activation function and provides a probability distribution within the five classes. Since accuracy is used as the assessment parameter, the model construction includes Adamax optimizer and categorical cross-entropy as a loss function. With the help of smart learning based on training data, such structured CNN enables accurate classification of images with features extracted.

## 3.4 End-End Proposed Framework for Rice Crop Disease Classification:

### 3.4.1 Image Acquisition:

For system development, the Rice Leaf Disease Images Dataset [77], which is openly accessible on Kaggle, is used. There are 6,000 photos in the suggested dataset, with 1500 photos for each type of disease.

### 3.4.2 Exploratory Data Analysis (EDA):

The data is split into three main parts *(as shown in Figure 3.12)*. Most of it - about 80% - is used to train the model. The rest is split evenly between testing and validation, with each getting around 10%. This is normal in machine learning - you use most of your data to train your model, then keep some aside to make sure it's working right and to test how well it performs with new information. This way of splitting up the data helps make sure your model learns properly and can handle new situations when you use it for real.

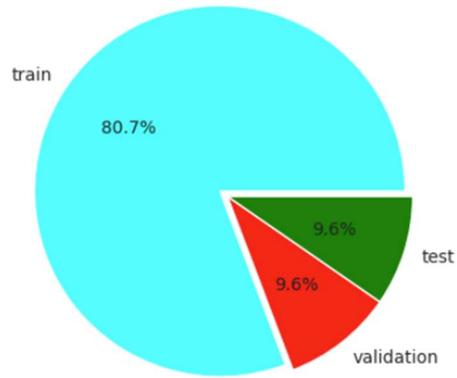

**Figure 3.12: Class Distribution of Different Varieties of Rice Grains**

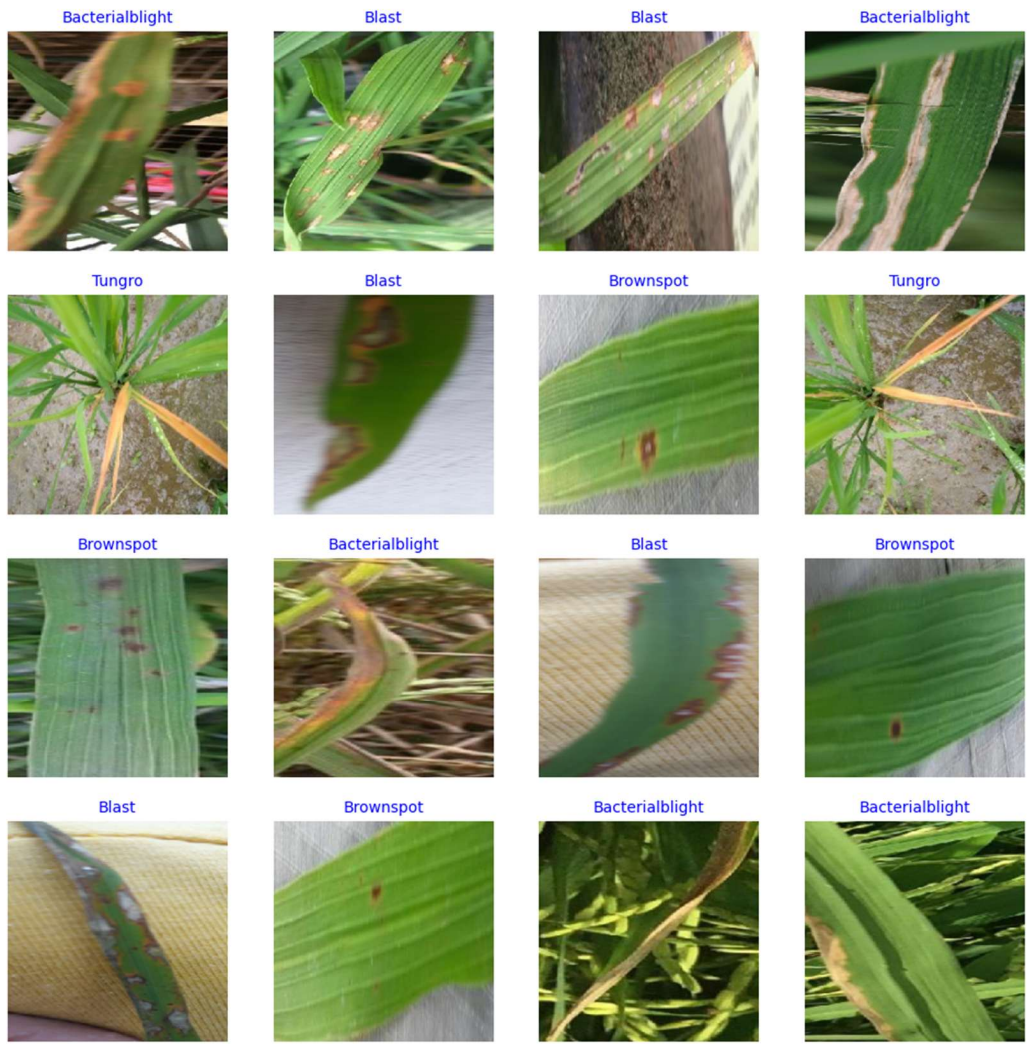

**Figure 3.13: Examples of Rice Crop Diseases: Bacterial Blight, Blast, Tungro, and Brown Spot as Identified in Samples**

### 3.4.3 Proposed Model:

For rice crop disease classification, a Convolutional Neural Network (CNN) has been carefully designed to accurately classify images into four different classes: Bacterial Blight, Blast, Tungro, and Brown Spot *(as shown in Figure 3.14)*. The health and yield of rice plants can be greatly impacted by the distinctive visual symptoms that each disease exhibits. This model aims to improve diagnostic accuracy and support efficient disease management techniques by utilizing deep learning.

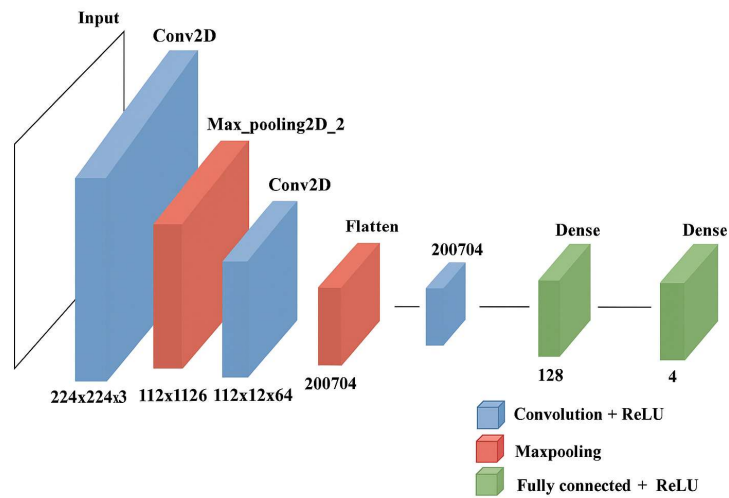

**Figure 3.14: Architecture of Proposed Model for Rice Crop Disease Classification**

The CNN architecture is designed for rice crop disease classification, utilizing state-of-the-art deep learning techniques to efficiently identify and classify common rice diseases. The model starts with an input layer that can process RGB-channel images with 224x224 pixels, ensuring a balance between computational efficiency and comprehensive disease identification. Multiple Convolutional Layers (Conv2D) are used in the first phase to extract features like edges, textures, and patterns from input images. The model uses 128 neurons in the first dense layer using the ReLU activation function and a SoftMax activation function in the last layer, which corresponds to the four disease classes.

To improve classification accuracy iteratively, training uses optimization techniques like Adam or SGD for weight adjustments based on computed gradients during backpropagation, along with a loss function like categorical cross-entropy to quantify prediction disparities. This methodical extraction and classification approach supports sustainable agriculture and food security by improving disease management techniques in rice cultivation and increasing diagnostic precision. The model begins with a convolutional layer (Conv2D) that uses 32 3x3 filters to learn spatial features like edges and textures. A ReLU activation function's non-linearity allows for the capture of intricate patterns. The architecture includes a max pooling layer (MaxPooling2D) to reduce the dimensions of feature maps and extract higher-level features. After pooling, a flattening layer transforms the multi-dimensional output into a one-dimensional vector for the fully connected layers.

# CHAPTER FOUR: CROP CLASSIFICATION EXPERIMENTAL RESULTS

## 4.1 Implementation Details:

Table 2 provides a comprehensive summary of the architecture and parameter specifics of the proposed convolutional neural network (CNN) model devised for rice grain classification. The model encompasses various layers, incorporating convolutional layers (Conv2D), max pooling layers (MaxPooling2D), culminating in a fully connected (Dense) layer. With a total of 267,397 trainable parameters, there are no frozen layers in the model. The input image size is estimated to be around 50×50×1, with the model tailored to classify images into five distinct output classes. This architecture underscores the model's complexity and capability to effectively learn from the rice grain dataset.

**Table 4.1: Architecture and Parameter Summary of the Proposed CNN Model for Rice Grain Classification.**

| Layer Name | Type | Output Shape | Parameters |
|---|---|---|---|
| conv2d_2 | Conv2D | (None, 48, 48, 32) | 896 |
| max_pooling2d_2 | MaxPooling2D | (None, 24, 24, 32) | 0 |
| conv2d_3 | Conv2D | (None, 22, 22, 64) | 18,496 |
| max_pooling2d_3 | MaxPooling2D | (None, 11, 11, 64) | 0 |
| flatten_1 | Flatten | (None, 7744) | 0 |
| dense_2 | Dense (F.C.) | (None, 32) | 247,840 |
| dense_3 | Dense (Output) | (None, 5) | 165 |
| Summary: | | | |
| Total Parameters: 267,397 | | | |
| Trainable Parameters: 267,397 | | | |
| Non-trainable Parameters: 0 | | | |
| Input Image Size (inferred): Likely 50×50×1 or 49×49×1 (based on backtracking conv/pooling layers) | | | |
| Output Classes: 5 | | | |

## 4.2 Training Details:

The tabulated data presents essential hyperparameters for the model, featuring a batch size of 32 and images formatted at 50x50 with 3 RGB channels. Utilizing the ADAM optimizer enhances model training efficiency, while incorporating L2 regularization aids in model regularization. These hyperparameter configurations are pivotal in fine-tuning the model's performance and facilitating successful training and classification of rice grain images.

Table 4.2: Hyperparameters Details

| Hyperparameter | Value |
|---|---|
| Batch Size | 32 |
| Image Size | 50x50 |
| No. of channels | 03 (RGB) |
| Optimizer | ADAM |
| Model Regularization | L2 Regularizartion |

## 4.3 Datasets:

### 4.3.1 Rice Images Dataset:

Table 4.3: Overall Description of Dataset

| Parameter Name | Description |
|---|---|
| Number of Classes | 05 |
| Names of Classes | Arborio, Basmati, Jasmine, Ipsala, and Karacadag |
| Total No. of Images | 75000 |
| Number of images per Class | 15000 |

## 4.4 Experimental Results:

Plots, *(as shown in Figure 4.1).*, provide important information about the convolutional neural networks (CNN) training dynamics, with an emphasis on accuracy and loss metrics throughout the training epochs. The training loss curve, shown in red on the left graph, shows a consistent drop over the course of training. With the loss getting closer to a lower bound, this steady decline shows that the model is successfully picking up patterns from the training set, indicating increased prediction accuracy. However, the validation loss, which is displayed in green, first decreases in a manner like that of the training loss, indicating that the model is initially successfully generalizing to the validation data.

However, as training goes on, the validation loss shows a slight increase, suggesting that overfitting may be starting. Overfitting occurs when the model starts to memorize the training data and loses its capacity to generalize new, unseen data. The **"best epoch"** point (at epoch 10) indicates the best time for model performance prior to this decline, implying that training should ideally be stopped at this point to preserve generalization abilities.

**Table 4.4: Detailed Classification Report of CNN Model for Different Rice Grain Types**

| Parameter Name | Precision | Recall | F1-Score |
|---|---|---|---|
| Arborio | 0.987 | 0.985 | 0.986 |
| Basmati | 0.993 | 0.989 | 0.991 |
| Ipsala | 0.998 | 1.00 | 0.997 |
| Jasmine | 0.986 | 0.991 | 0.989 |
| Karacadag | 0.987 | 0.992 | 0.990 |

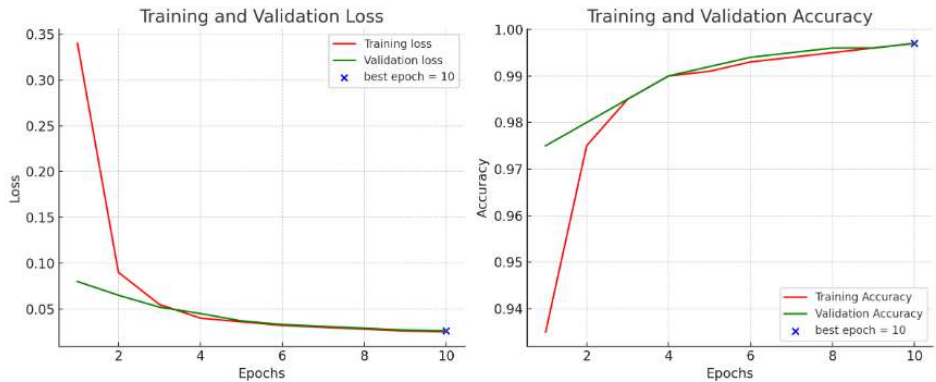

**Figure 4.1: Training loss and accuracy curves illustrating model performance over epochs**

A thorough summary of the convolutional neural network's classification performance across five different classes—Arborio, Basmati, Ipsala, Jasmine, and Karacadag—is given by the confusion matrix *(as shown in Figure 4.2).* The true labels are represented by each row of the matrix, and the model's predicted labels are represented by each column. The model's accuracy in identifying each rice variety is demonstrated by the diagonal entries, which show the instances for each class that were correctly classified. For example, the model demonstrated high accuracy for the classes Arborio, Basmati, and Ipsala by correctly classifying 1,478 instances, 1,483 instances, and 1,495 instances, respectively.

The areas where our suggested model struggles, especially when it comes to incorrectly classifying the instances in off-diagonal cells, are revealed by the confusion matrix (as illustrated in Fig. 10). For instance, it shows that 10 cases of jasmine grains are misclassified as karacadag, 4 cases of arborio were misclassified as ipsala, and 16 cases of ipsala were misclassified as basmati. These results are crucial for drawing attention to the model's shortcomings, particularly when it comes to differentiating between two closely related classes.

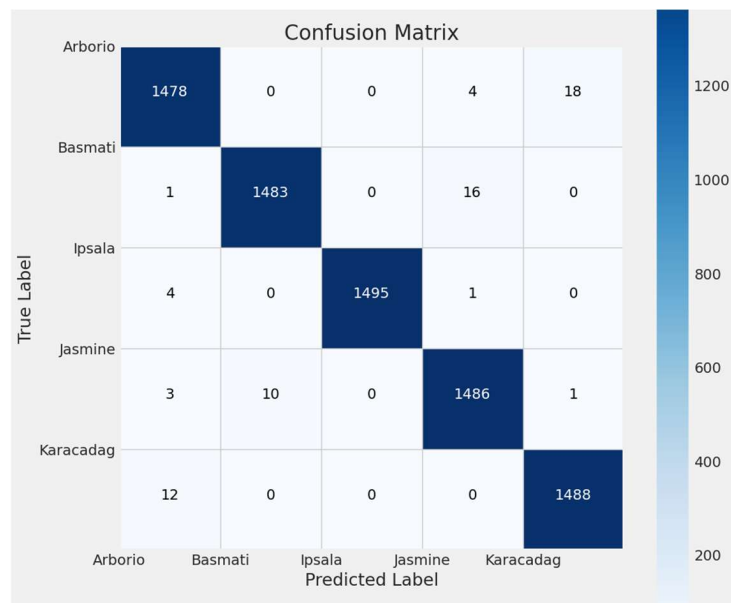

**Figure 4.2: Confusion matrix displaying classification results and model performance**

As the input for the classification model, the image *(as shown in Figure 4.3),* the *"Original Image",* shows a grain of rice. Training the model to identify and distinguish between different types of rice based on their visual characteristics requires this clear representation. The model's capacity to learn and generate precise predictions is directly impacted by the caliber and clarity of the input images.

The image *(as shown in Figure 4.3)* titled *"LIME Explanation"* displays the outcomes of the CNN's predictions using the *LIME (Local Interpretable Model-agnostic Explanations)* technique. The areas of the rice grain that the model determined to be most significant in its classification decision are indicated here by the yellow outline. In addition to improving the model's interpretability, this visualization sheds light on the characteristics that influence its predictions. LIME helps stakeholders and researchers comprehend the logic behind the model by highlighting the important aspects of the image, which promotes confidence in the model's results. All things considered, these pictures effectively convey the two main goals of our study: attaining high classification accuracy and guaranteeing openness in the machine learning models' decision-making process.

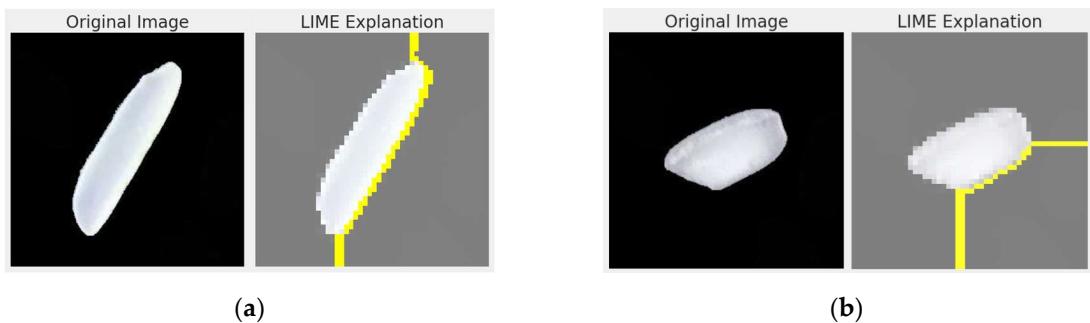

(a)          (b)

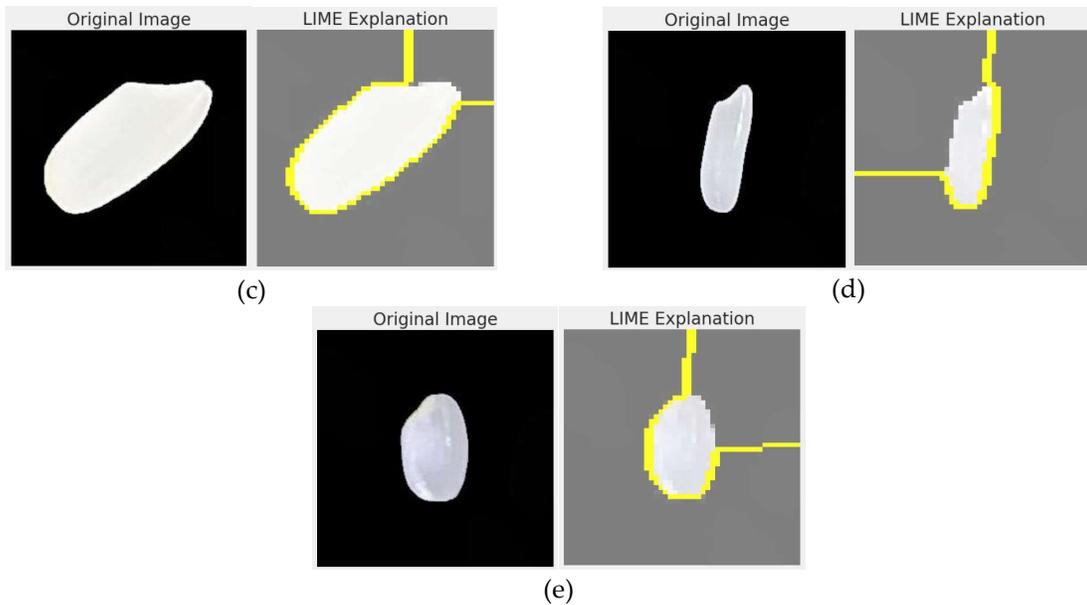

(c)                 (d)

(e)

**Figure 4.3: Original Image vs. LIME-generated visualization highlighting influential features (a) Basmati Rice Grain, (b) Arborio Rice Grain, (c) Ipsala Rice Grain, (d) Jasmine Rice Grain, (e) Karacadag Rice Grain**

Regarding the classification of Basmati rice grain across multiple outputs in the form of output 0\~3, the SHAP analysis illustrated in the image *(as shown in Figure 4.4)* offers important insights. Every output denotes a distinct class associated with rice variety, and the corresponding SHAP values show how each pixel contributed to the model's predictions.

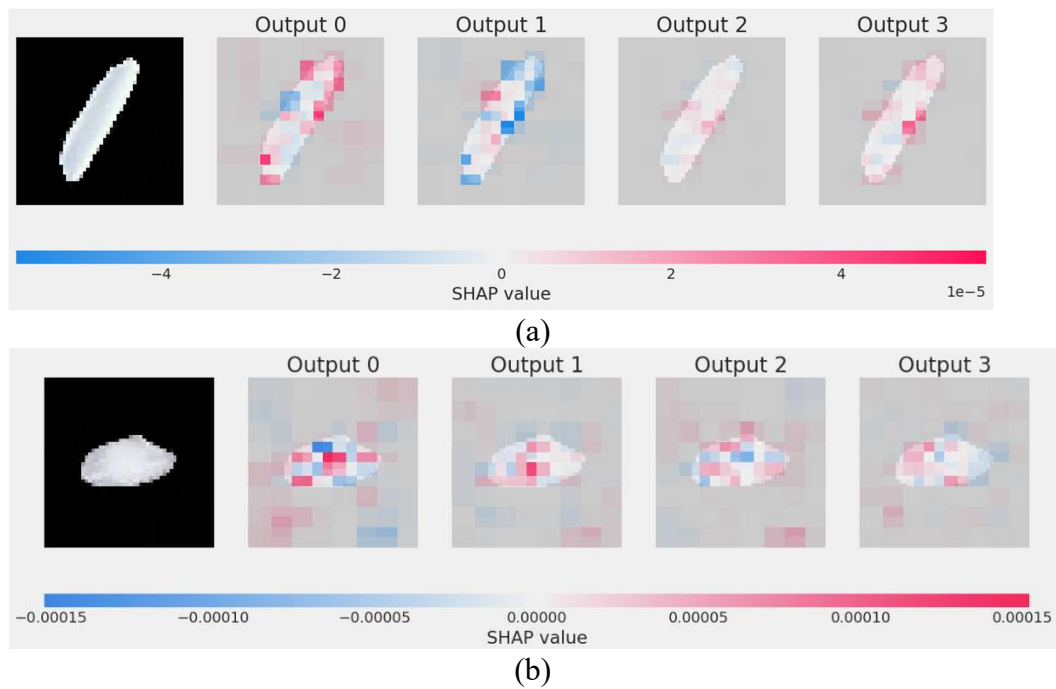

(a)

(b)

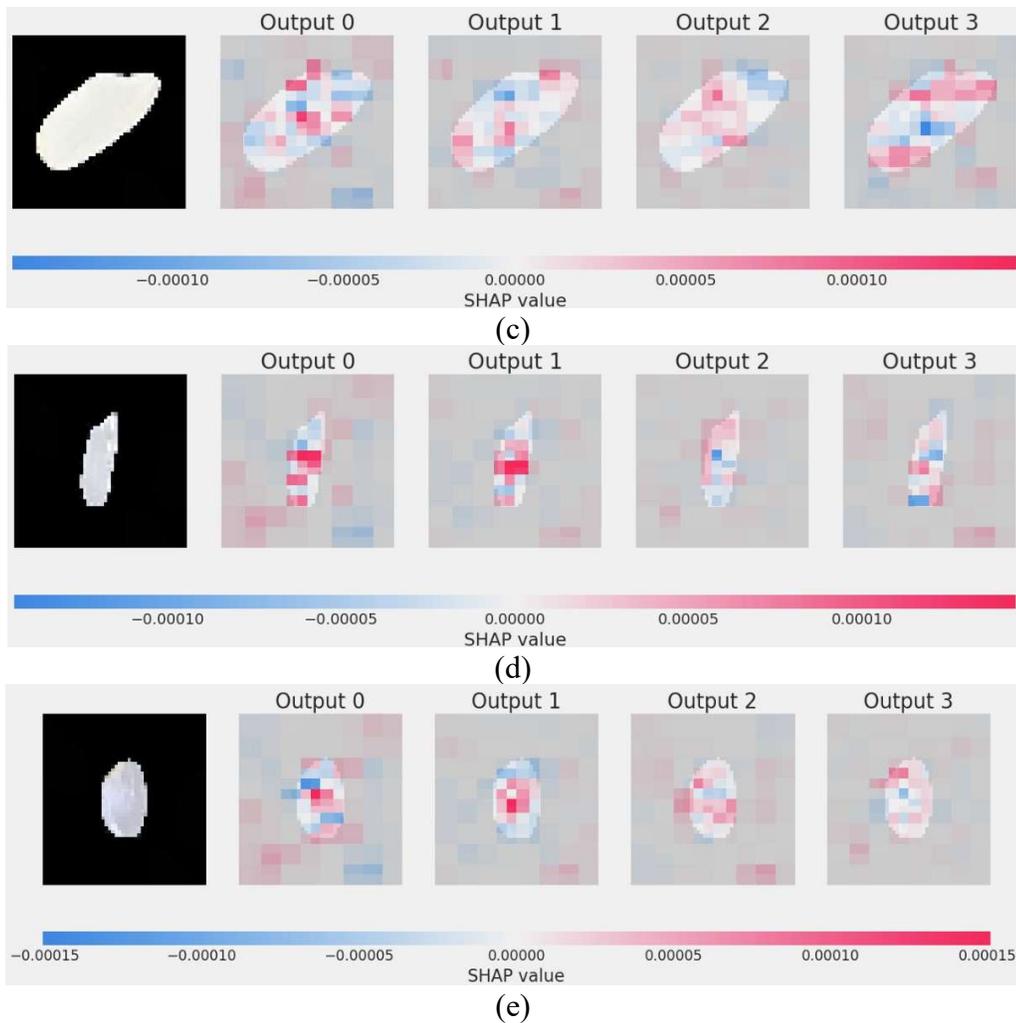

(c)

(d)

(e)

**Figure 4.4: SHAP-generated output highlights the key features; (a) Basmati rice grain, (b) Arborio Rice Grain, (c) Ipsala Rice Grain, (d) Jasmine Rice Grain, (e) Karacadag Rice Grain**

# CHAPTER FIVE: CROP DISEASE CLASSIFICATION EXPERIMENTAL RESULTS

## 5.1 Implementation Details:

### 5.1.1 Network Architecture:

### 5.1.2 Training Details:

Table 5.1: Hyperparameters Details

| Hyperparameter | Value |
|---|---|
| Batch Size | 32 |
| Image Size | 50x50 |
| No. of channels | 03 (RGB) |
| Optimizer | ADAM |
| Model Regularization | L2 Regularizartion |

## 5.2 Datasets:

### 5.2.1 Rice Images Dataset:

Table 5.2: Overall Description of Dataset

| Parameter Name | Description |
|---|---|
| Number of Classes | 04 |
| Names of Classes | Bacterial Blight, Burst, Brown Spot, Tungro |
| Total No. of Images | 6000 |
| Number of images per Class | 1500 |

## 5.3 Experimental Results:

The graphs *(as shown in Figure 5.1)* depict the progression of training and validation loss and accuracy across 80 epochs for the rice crop disease classification model. In the left chart, the training loss (in red) steadily decreases throughout epochs, indicating effective learning from the training data. Validation loss (in green) also shows a decreasing trend but with some fluctuations, implying overall improvement with occasional performance variability on the validation set. The point of lowest validation loss highlights the optimal epoch, underscoring the importance of monitoring validation loss to prevent overfitting and ensure generalization to new data.

On the right chart, training accuracy (red) shows a consistent rise, signifying successful learning of training data characteristics. Validation accuracy (green) increases as well, albeit at a slightly slower pace, revealing a performance gap between training and validation sets. The convergence of both accuracy curves towards high values suggests positive model progress, although continuous monitoring is crucial for refining performance. These metrics collectively offer valuable insights into the model's learning dynamics and efficacy in classifying rice crop diseases.

**Table 5.3:** Detailed Classification Report of CNN Model for Rice Crop Disease Classification

| Parameter Name | Precision | Recall | F1-Score |
|---|---|---|---|
| Bacterial Blight | 1.00 | 0.97 | 0.99 |
| Blast | 0.95 | 1.00 | 0.98 |
| Brown Spot | 1.00 | 0.98 | 0.99 |
| Tungro | 0.99 | 0.99 | 0.99 |

**Table 5.4:** Performance Comparison of Pretrained CNN Models

| Model | Accuracy | Loss |
|---|---|---|
| **RESNET-50** | 99.82% | 13.47% |
| **VGG-16** | 86.7% | 75.95% |
| **MobileNet-V2** | 77.3% | 55.42% |
| **DenseNet121** | 90.07% | 26.11% |

This table presents the classification accuracy and loss values of four pretrained convolutional neural network architectures (ResNet-50, VGG-16, MobileNet-V2, and DenseNet121) evaluated on the same dataset. ResNet-50 outperformed all other models with the highest accuracy (99.82%) and the lowest loss (13.47%).

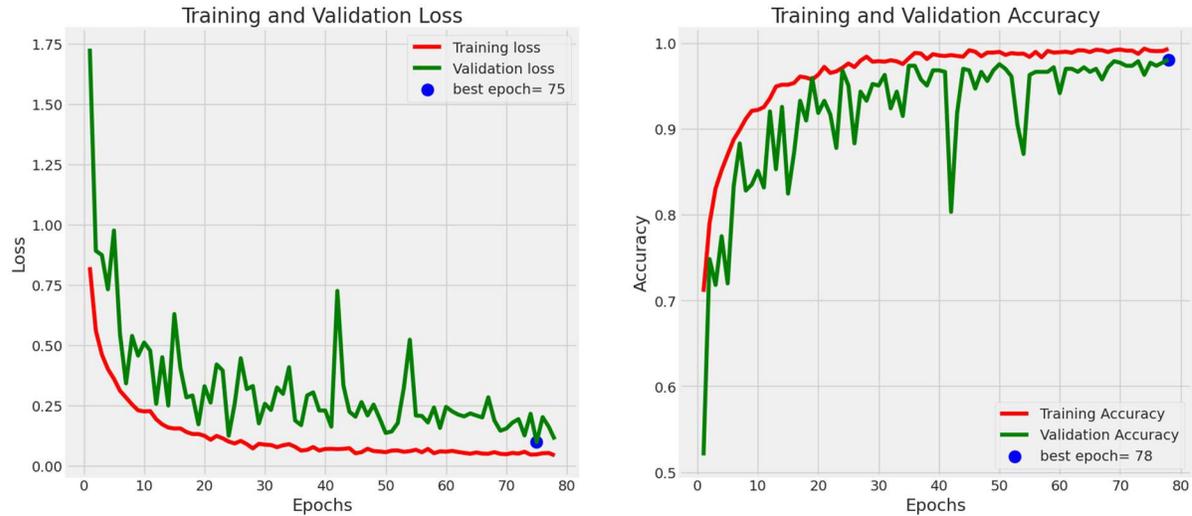

**Figure 5.1: Training loss and accuracy curves illustrating model performance over epochs**

The confusion matrix *(as shown in Figure 5.2)* offers a detailed overview of the classification performance of the rice crop disease model concerning the four target classes: Bacterial Blight, Blast, Brown Spot, and Tungro. Rows in the matrix represent true labels, while columns depict predicted labels. Correct classifications are reflected in diagonal entries, showcasing high accuracy for Bacterial Blight (147), Brown Spot (149), and Tungro (123), underscoring the model's proficiency in identifying these diseases. Despite this success, some misclassifications are evident, notably with Blast, where three instances were misclassified as Bacterial Blight and one as Tungro. The confusion matrix highlights the model's strengths in precise disease diagnosis while pinpointing areas for enhancement, particularly in distinguishing closely related classes. This analysis is pivotal for refining the model and elevating its diagnostic capabilities in practical agricultural contexts.

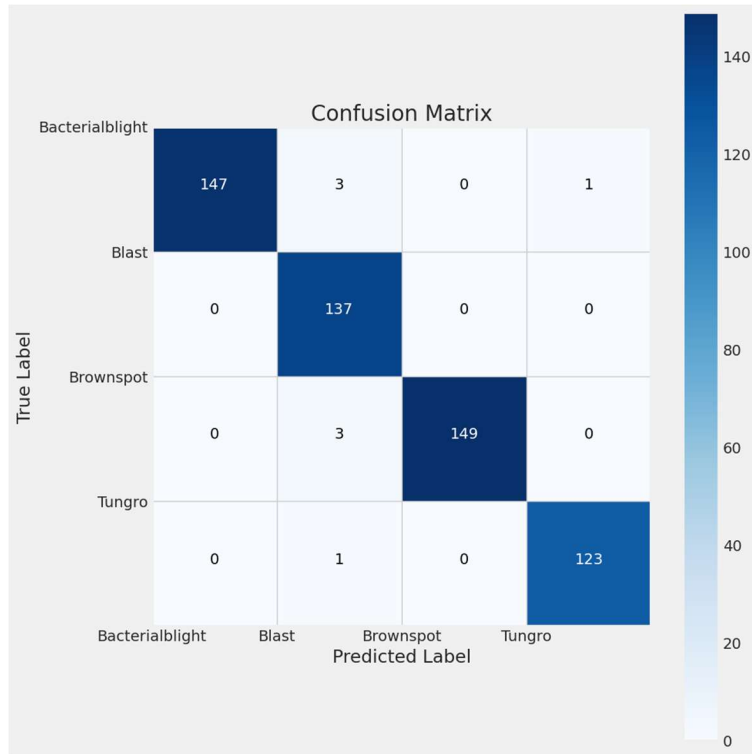

**Figure 5.2: Confusion matrix displaying classification results and model performance for Rice Diseases**

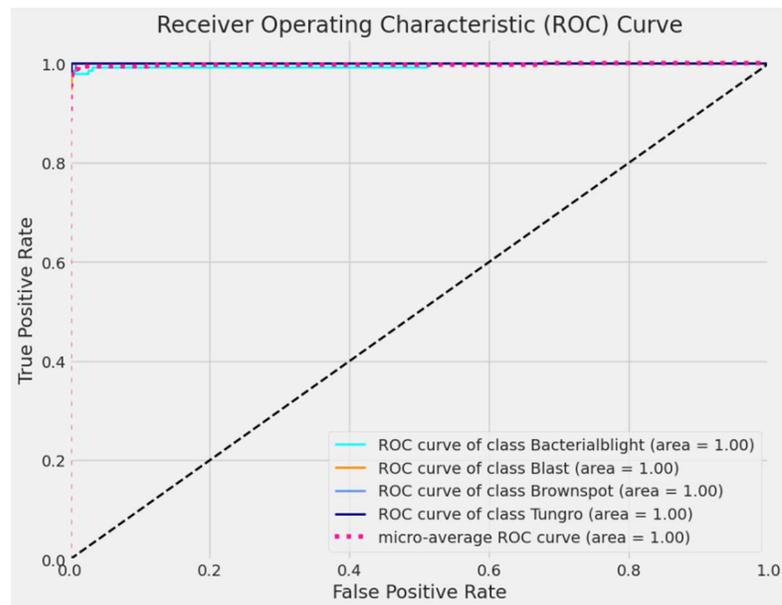

**Figure 5.3:RoC for Crop disease Classification**

ROC curve shows the accuracy at which the model has classified Bacterial blight, Blast, Brownspot, and Tungro with a micro-average ROC curve. The fact that an AUC of 1.00 is reached across the boundaries is evidence of the high-performance standards of the model on different thresholds as well.

As shown in Fig. 12(a), the input of the disease classification model is called the Original Image which shows the visual symptoms of rice leaf. A clear and quality image is needed as an input into the model so that it can be trained to be able to recognise and classify different rice leaf diseases such as blast, bacterial blight, brown spot, tungro among others based on patterns recognizable on the leaf such as discolouration, lesions, etc. These clearly defined images in these models have significant influence in the ability of the model to define discriminative features and make perfect predictions.

The image (shown in part (a)), also called the Original Image, is inserted to the disease classification model and represents a rice leaf with visual symptoms. The input should be a good quality and clear picture that will help model be trained to distinguish and categorize between different rice leaf diseases such as blast, bacterial blight, brown spot and tungro in terms of evident patterns such as lesions, discoloration and spots. These clearly defined images in these models have significant influence in the ability of the model to define discriminative features and make perfect predictions.

The image (see part (b) is called the LIME Explanation, which demonstrates the interpretation of the input of the model based on the LIME (Local Interpretable Model-agnostic Explanations) approach. The colored overlays or contours, commonly known as the highlighted regions, denote the areas of the leaf that the model found most effective in coming up with its decision about the disease upon inspection of the leaf. This makes the model based on the CNN more explainable, but it also allows the agronomists and the researchers to see the motivations behind the predictions made by the model. LIME creates a level of transparency with the generation of visual representations of the main characteristics the model is based on, thus instilling confidence into the AI-aided diagnosis, and, eventually, contributing to timely and informed plant disease management decisions.

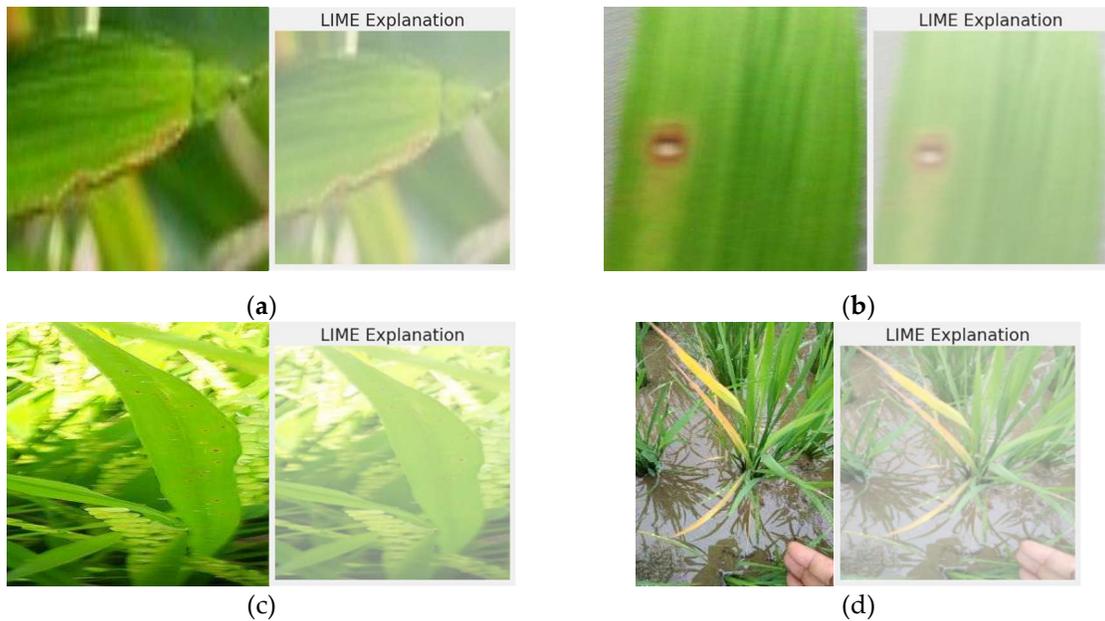

**Figure 5.4:Original Image vs. LIME-generated visualization using MobileNet-V2 highlighting influential features (a) Leaves with Bacterial blight Disease, (b) Leaves with Blast Disease, (c) Leaves with Brown Spot Disease, (d) Leaves with Tungro Disease**

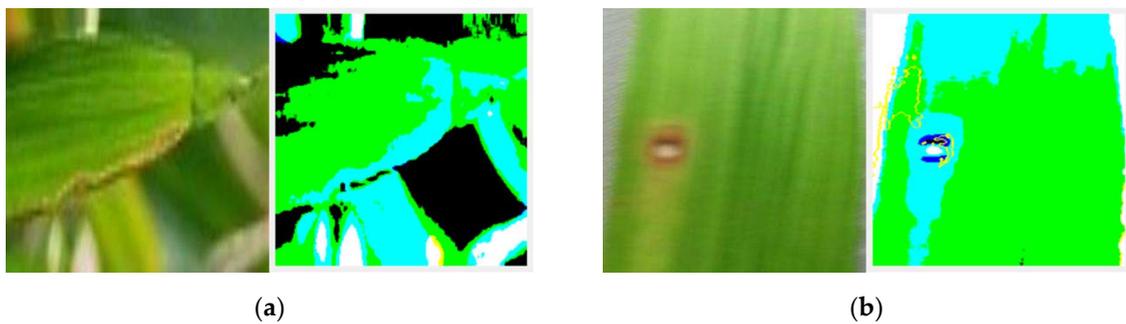

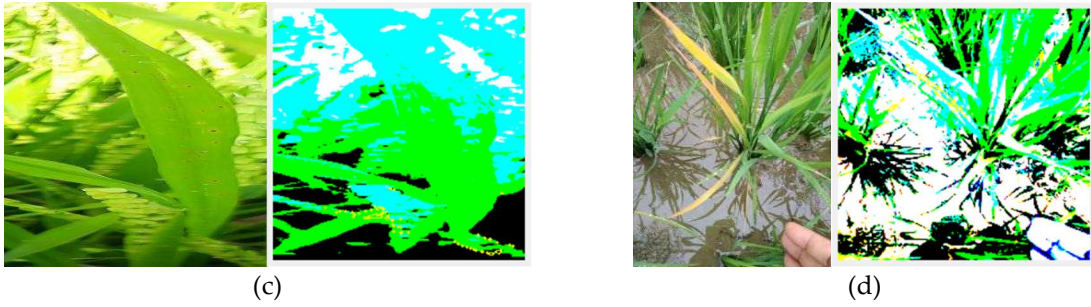

(c)                                              (d)

**Figure 5.5:** Original Image vs. LIME-generated visualization using VGG16 highlighting influential features **(a)** Leaves with Bacterial Blight Disease, **(b)** Leaves with Blast Disease, **(c)** Leaves with Brown Spot Disease, **(d)** Leaves with Tungro Disease

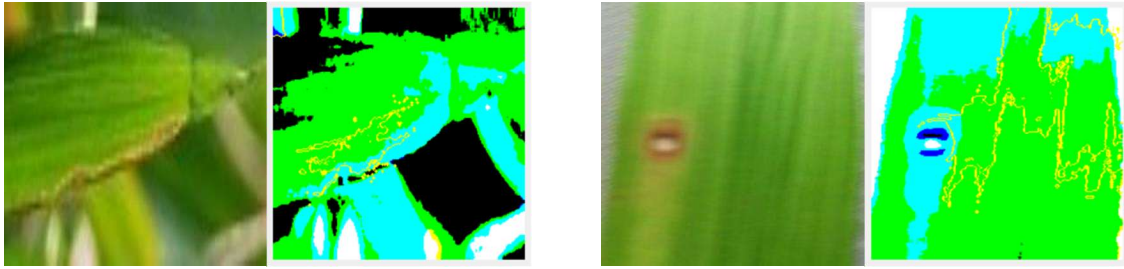

(a)                                              (b)

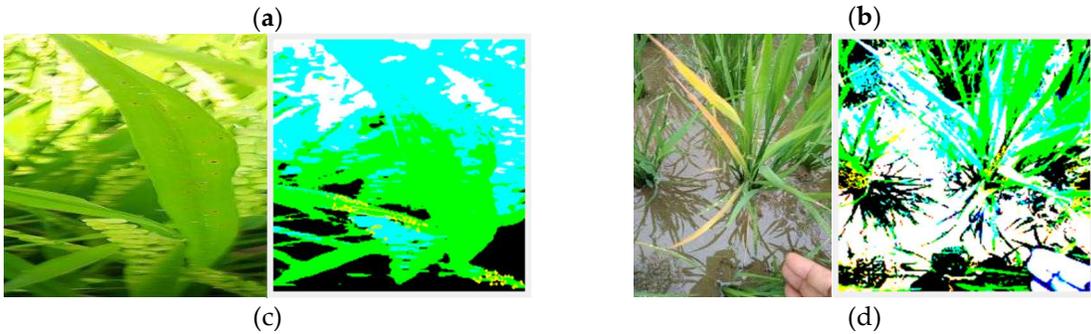

(c)                                              (d)

**Figure 5.6:** Original Image vs. LIME-generated visualization using RESNET-50 highlighting influential features **(a)** Leaves with Bacterial Blight Disease, **(b)** Leaves with Blast Disease, **(c)** Leaves with Brown Spot Disease, **(d)** Leaves with Tungro Disease

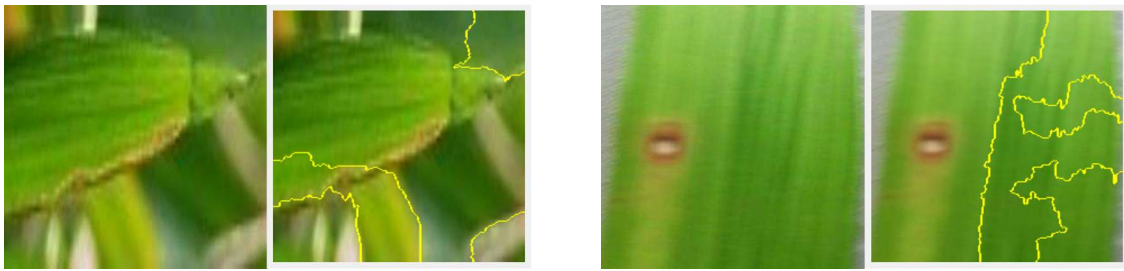

(a)                                              (b)

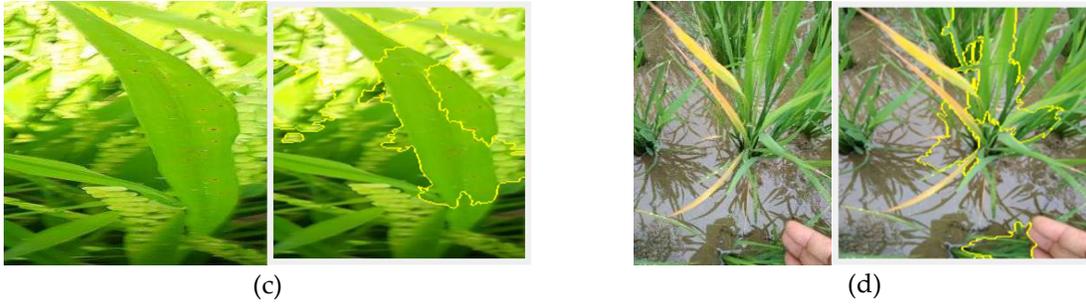

(c)                                  (d)

**Figure 5.7: Original Image vs. LIME-generated visualization using MobileNet-V2 highlighting influential features (a) Leaves with Bacterial blight Disease, (b) Leaves with Blast Disease, (c) Leaves with Brown Spot Disease, (d) Leaves with Tungro Disease**

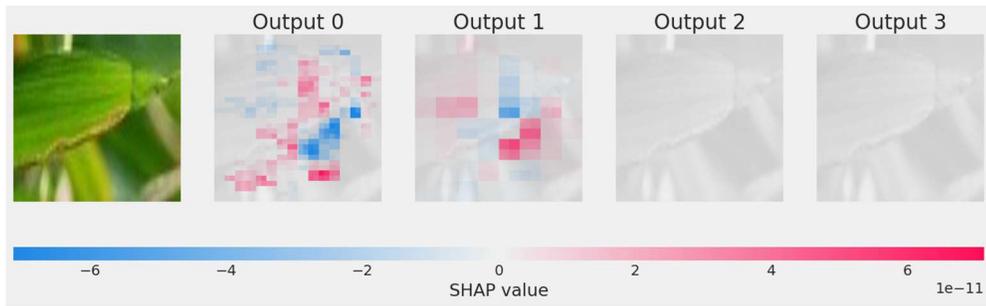

(a)

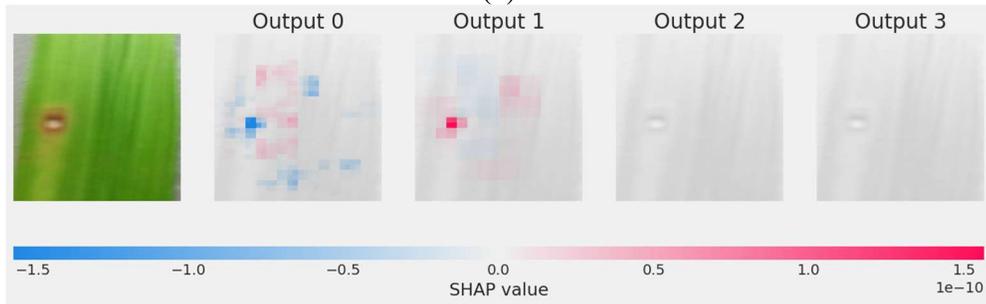

(b)

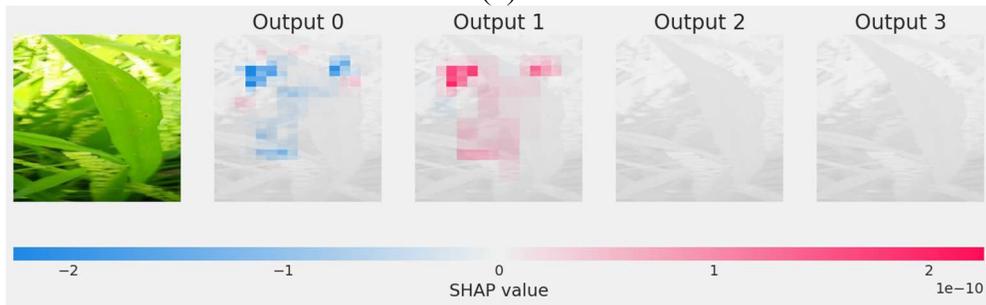

(c)

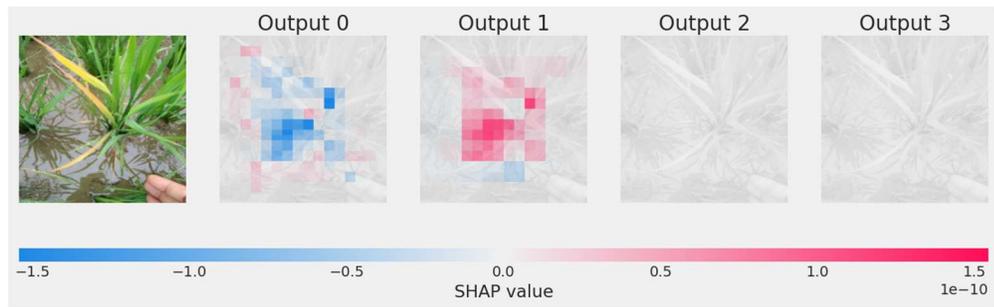

(d)

**Figure 5.8:** SHAP-generated output using Custom CNN highlights the key features; (a) Leaves with Bacterial Blight Disease, (b) Leaves with Blast Disease, (c) Leaves with Brown Spot Disease, (d) Leaves with Tungro Disease

# CHAPTER SIX: DISCUSSIONS

Following the concepts of rice grain classification and detection of rice crops disease, the thesis titled Advancements in Crop Analysis through Deep Learning and Explainable AI provides a comprehensive schema of crop analysis. As the means of enhancing the accuracy of the models and making them more comprehensible to a user, the proposed architecture includes explainable AI tools, such as the SHAP and LIME with computer vision and deep learning as the latest technologies.

In order to enhance the supply chain, market segment differentiation and quality assurance, the proposed method utilizes an automated mechanism of grain classification of different varieties of rice grains. To enhance the resilience of crops and promote environmentally sustainable agricultural procedures, an early and accurate technique of detecting rice leaf diseases including Blast, Bacteria Blight, Brown Spot, and Tungro can be also formed.

Experimental outcomes when determining the crop type in the phase of classification of crops attest the potential of deep learning models (i.e. CNN, VGG16 and RESNET-50) to characterize with a high probability the various types of rice. Their high percentages of accuracy rates reveal their effectiveness in real practice of the models. The use of explainable AI approaches like SHAP and LIME makes the models easier to use and understand since they provide information on how the models came to their decision.

A variety of deep learning models is tested in terms of sensitivity and precision in crop disease detection experiments concerning the topping of the most common rice leaf diseases. The results prove the effectiveness of continuous monitoring and early identification of diseases in agriculture, since the two models like InceptionResNetV2 and DenseNet201 have shown that they are better in reaching high levels of accuracies.

Modeling of the stains depends on the explainability component, which is crucial in comparing the machine and deep learning processes in the detection and classification of rice leaf diseases. The models will also be more trustworthy and useable since they explain the rationale of the forecasts through tools such as LIME and SHAP, which enables the stakeholders to make sound crop management decisions.

All in all, the thesis will become a valuable contribution to the sphere of crop analysis due to the proclaimed development of new advanced models, in-depth experiments development, and evidence. This can utilize the combination of deep learning, computer vision, and explainable AI to address these issues by improved crop classification accuracy and disease detection performance in order to promote sustainable agricultural production modes and guarantee food security by region.

[52] H. Khalid, S. Saqib, M. J. Asif and D. A. Dewi, "Strategic Customer Segmentation: Harnessing Machine Learning For Retaining Satisfied Customers," *Lahore Garrison University Research Journal of Computer Science and Information Technology,* vol. 8, 2024.

[53] M. Abouelyazid, "Advanced artificial intelligence techniques for real-time predictive maintenance in industrial IoT systems: a comprehensive analysis and framework," *J. AI-Assist. Sci. Discov,* vol. 3, p. 271–313, 2023.

[54] B. Abhisheka, S. K. Biswas and B. Purkayastha, "A comprehensive review on breast cancer detection, classification and segmentation using deep learning," *Archives of Computational Methods in Engineering,* vol. 30, p. 5023–5052, 2023.

[55] H. Jiang, Z. Diao, T. Shi, Y. Zhou, F. Wang, W. Hu, X. Zhu, S. Luo, G. Tong and Y.-D. Yao, "A review of deep learning-based multiple-lesion recognition from medical images: classification, detection and segmentation," *Computers in Biology and Medicine,* vol. 157, p. 106726, 2023.

[56] I. Pacal and S. Kılıcarslan, "Deep learning-based approaches for robust classification of cervical cancer," *Neural Computing and Applications,* vol. 35, p. 18813–18828, 2023.

[57] B. S. Abunasser, M. R. J. Al-Hiealy, I. S. Zaqout and S. S. Abu-Naser, "Convolution neural network for breast cancer detection and classification using deep learning," *Asian Pacific journal of cancer prevention: APJCP,* vol. 24, p. 531, 2023.

[58] S. Ruksakulpiwat, W. Thongking, W. Zhou, C. Benjasirisan, L. Phianhasin, N. K. Schiltz and S. Brahmbhatt, "Machine learning-based patient classification system for adults with stroke: a systematic review," *Chronic Illness,* vol. 19, p. 26–39, 2023.

[59] D.-h. Kim, K. Oh, S.-h. Kang and Y. Lee, "Development of Pneumonia Patient Classification Model Using Fair Federated Learning," in *International Conference on Intelligent Human Computer Interaction*, 2023.

[60] M. Koklu, I. Cinar and Y. S. Taspinar, "Classification of rice varieties with deep learning methods," *Computers and electronics in agriculture,* vol. 187, p. 106285, 2021.

[61] F. D. Adhinata, R. Sumiharto and others, "A comprehensive survey on weed and crop classification using machine learning and deep learning," *Artificial intelligence in agriculture,* 2024.

[62] D. Agarwal, P. Bachan and others, "Machine learning approach for the classification of wheat grains," *Smart Agricultural Technology,* vol. 3, p. 100136, 2023.

[63] Y. Ramdhani and D. P. Alamsyah, "Enhancing Sustainable Rice Grain Quality Analysis with Efficient SVM Optimization Using Genetic Algorithm," in *E3S Web of Conferences*, 2023.

[64] G. Sakkarvarthi, G. W. Sathianesan, V. S. Murugan, A. J. Reddy, P. Jayagopal and M. Elsisi, "Detection and classification of tomato crop disease using convolutional neural network," *Electronics,* vol. 11, p. 3618, 2022.
70

[92] O. Rama Devi, M. S. Al Ansari, B. P. Reddy, M. HT, Y. A. Baker El-Ebiary, M. Rengarajan and others, "Optimizing Crop Yield Prediction in Precision Agriculture with Hyperspectral Imaging-Unmixing and Deep Learning.," *International Journal of Advanced Computer Science & Applications,* vol. 14, 2023.

[93] S. Qadri, T. Aslam, S. A. Nawaz, N. Saher, A. Razzaq, M. Ur Rehman, N. Ahmad, F. Shahzad and S. Furqan Qadri, "Machine vision approach for classification of rice varieties using texture features," *International Journal of Food Properties,* vol. 24, p. 1615–1630, 2021.

[94] P. K. Priya, P. Kirupa, P. Thilakaveni, K. N. Devi, M. Mahabooba and S. Jayachitra, "DeepRiceTransfer: Exploiting CNN Transfer Learning for Effective Rice Variety Classification," in *2024 International Conference on Social and Sustainable Innovations in Technology and Engineering (SASI-ITE)*, 2024.

[95] K. A. Patil and N. R. Kale, "A model for smart agriculture using IoT," in *2016 international conference on global trends in signal processing, information computing and communication (ICGTSPICC)*, 2016.

[96] I. M. Nasir, A. Bibi, J. H. Shah, M. A. Khan, M. Sharif, K. Iqbal, Y. Nam and S. Kadry, "Deep learning-based classification of fruit diseases: An application for precision agriculture," 2021.

[97] I. S. Na, S. Lee, A. M. Alamri and S. A. AlQahtani, "Remote sensing and AI-based monitoring of legume crop health and growth," *Legume Research,* vol. 47, p. 1179–1184, 2024.

[98] A. Monteiro, S. Santos and P. Gonçalves, "Precision agriculture for crop and livestock farming—Brief review," *Animals,* vol. 11, p. 2345, 2021.

[99] C. Lytridis, V. G. Kaburlasos, T. Pachidis, M. Manios, E. Vrochidou, T. Kalampokas and S. Chatzistamatis, "An overview of cooperative robotics in agriculture," *Agronomy,* vol. 11, p. 1818, 2021.

[100] S. Kujawa and G. Niedbała, *Artificial neural networks in agriculture,* vol. 11, MDPI, 2021, p. 497.

[101] S. Khanal, K. Kc, J. P. Fulton, S. Shearer and E. Ozkan, "Remote sensing in agriculture—accomplishments, limitations, and opportunities," *Remote sensing,* vol. 12, p. 3783, 2020.

[102] M. M. Islam, M. A. A. Adil, M. A. Talukder, M. K. U. Ahamed, M. A. Uddin, M. K. Hasan, S. Sharmin, M. M. Rahman and S. K. Debnath, "DeepCrop: Deep learning-based crop disease prediction with web application," *Journal of Agriculture and Food Research,* vol. 14, p. 100764, 2023.

[103] S. Ibrahim, N. A. Zulkifli, N. Sabri, A. A. Shari and M. R. M. Noordin, "Rice grain classification using multi-class support vector machine (SVM)," *IAES International Journal of Artificial Intelligence,* vol. 8, p. 215, 2019.

[104] D. M. K. S. Hemathilake and D. M. C. C. Gunathilake, "Agricultural productivity and food supply to meet increased demands," in *Future foods*, Elsevier, 2022, p. 539–553.
73